%% file: cam_ready.tex
\documentclass{article}

\usepackage[final,main]{neurips_2026}

\usepackage[utf8]{inputenc}
\usepackage[T1]{fontenc}
\usepackage{url}
\usepackage{nicefrac}
\usepackage{microtype}
\usepackage{graphicx}
\usepackage{subcaption}
\usepackage{wrapfig}
\usepackage{booktabs}
\usepackage{hyperref}
\usepackage{amsfonts}

\newcommand{\ALGCOMMENT}[1]{\hfill\(\triangleright\)~\textit{#1}}

\usepackage{amsmath}
\usepackage{amssymb}
\usepackage{mathtools}
\usepackage{amsthm}
\usepackage[capitalize,noabbrev]{cleveref}

\usepackage{bm}
\usepackage{algorithm}
\usepackage{algorithmic}
\usepackage[dvipsnames]{xcolor}
\usepackage{tikz}
\usepackage[most]{tcolorbox}
\usepackage{enumitem}
\usepackage{dblfloatfix}
\setlist[enumerate]{nosep,leftmargin=*,itemsep=1pt,topsep=2pt,parsep=0pt,partopsep=0pt}
\setlist[itemize]{nosep,leftmargin=*,itemsep=1pt,topsep=2pt,parsep=0pt,partopsep=0pt}

\input{DL_notation}

\newcommand{\beq}{\begin{equation}}
\newcommand{\eeq}{\end{equation}}
\newcommand{\be}{\begin{equation}}
\newcommand{\ee}{\end{equation}}
\newcommand{\beqa}{\begin{eqnarray}}
\newcommand{\eeqa}{\end{eqnarray}}
\newcommand{\bean}{\begin{eqnarray*}}
\newcommand{\eean}{\end{eqnarray*}}

\usepackage{tcolorbox}
\definecolor{lightblue}{HTML}{ff7f2a}
\definecolor{lighterblue}{HTML}{ffe6d5}
\newtcolorbox{summarybox}{
  colback=lighterblue,
  colframe=lightblue,
  breakable,       
  enhanced jigsaw, 
  before skip=10pt,
  after skip=10pt
}

\theoremstyle{plain}
\newtheorem{theorem}{Theorem}[section]
\newtheorem{lemma}{Lemma}[section]
\theoremstyle{definition}
\newtheorem{definition}[theorem]{Definition}

\newcommand{\cH}{\mathcal{H}}
\newcommand{\llqrdiag}{\textcolor{blue}{LLQR}}
\newcommand{\llqrkfac}{\textcolor{green!60!black}{LLQR}}
\newcommand{\newton}{\textsuperscript{$\circ$}}
\newcommand{\ngd}{\textsuperscript{$\triangle$}}

\newcommand{\crcolor}{black}
\newcommand{\crtext}[1]{{\color{\crcolor}#1}}

\newcommand{\bluetext}[1]{\textcolor{black}{#1}}

\newcommand{\mdfy}[1]{\textcolor{black}{#1}}

\title{Layerwise LQR for Geometry-Aware Optimization of Deep Networks}

\author{%
  \crtext{Simon Dufort-Labbé\thanks{\crtext{Corresponding author: \texttt{simon.dufort-labbe@mila.quebec}.}}}\\
  \crtext{Mila, Université de Montréal}\\
  \And
  \crtext{Pierre-Luc Bacon}\\
  \crtext{Canada CIFAR AI Chair}\\
  \crtext{Mila, Université de Montréal}
  \And
  \crtext{Razvan Pascanu}\\
  \crtext{Mila, Université de Montréal}
  \And
  \crtext{Simon Lacoste-Julien}\\
  \crtext{Canada CIFAR AI Chair}\\
  \crtext{Samsung -- SAIL Montreal}\\
  \crtext{Mila, Université de Montréal}\\
  \And
  \crtext{Aristide Baratin}\\
  \crtext{Samsung -- SAIL Montreal}\\
  \crtext{Mila, Université de Montréal}\\
}

\begin{document}

\maketitle

\begin{abstract}
    Geometry-aware optimizers such as Newton and natural gradient can improve conditioning in deep learning, but scalable variants such as K-FAC, Shampoo, and related preconditioners usually impose structural approximations early, often discarding cross-layer interactions induced by the network computation. We introduce Layerwise LQR (LLQR), a framework for learning structured inverse preconditioners under a global layerwise optimal-control objective.  The starting point is an exact equivalence: the steepest-descent step under a broad class of divergence-induced quadratic models—including Newton, Gauss–Newton, Fisher/natural-gradient, and intermediate-layer metrics—can be written as a finite-horizon Linear Quadratic Regulator (LQR) problem. 
    This formulation serves as a reference that exposes the layerwise dynamics and cost matrices encoding the original dense geometry. We then derive a scalable relaxation that learns diagonal, (E-)Kronecker-factored, or other structured inverse preconditioners by minimizing the LQR objective and reusing them across iterations. The resulting optimizer wraps standard methods while retaining a principled connection to second-order geometry, without forming or inverting the global curvature matrix. 
    Experiments 
    on ResNets and Transformers 
    show that LLQR
    improves optimization dynamics and often translates these gains into improved final test performance, while adding only modest wall-clock overhead.  It establishes LLQR as a practical framework for geometry-aware second-order methods and a reference for evaluating scalable approximations.
    \crtext{Implementation is available at: \href{https://github.com/SimonDufLab/LLQR}{github.com/SimonDufLab/LLQR}}
\end{abstract}

\begin{figure*}[t]
  \centering
  \includegraphics[width=\textwidth]{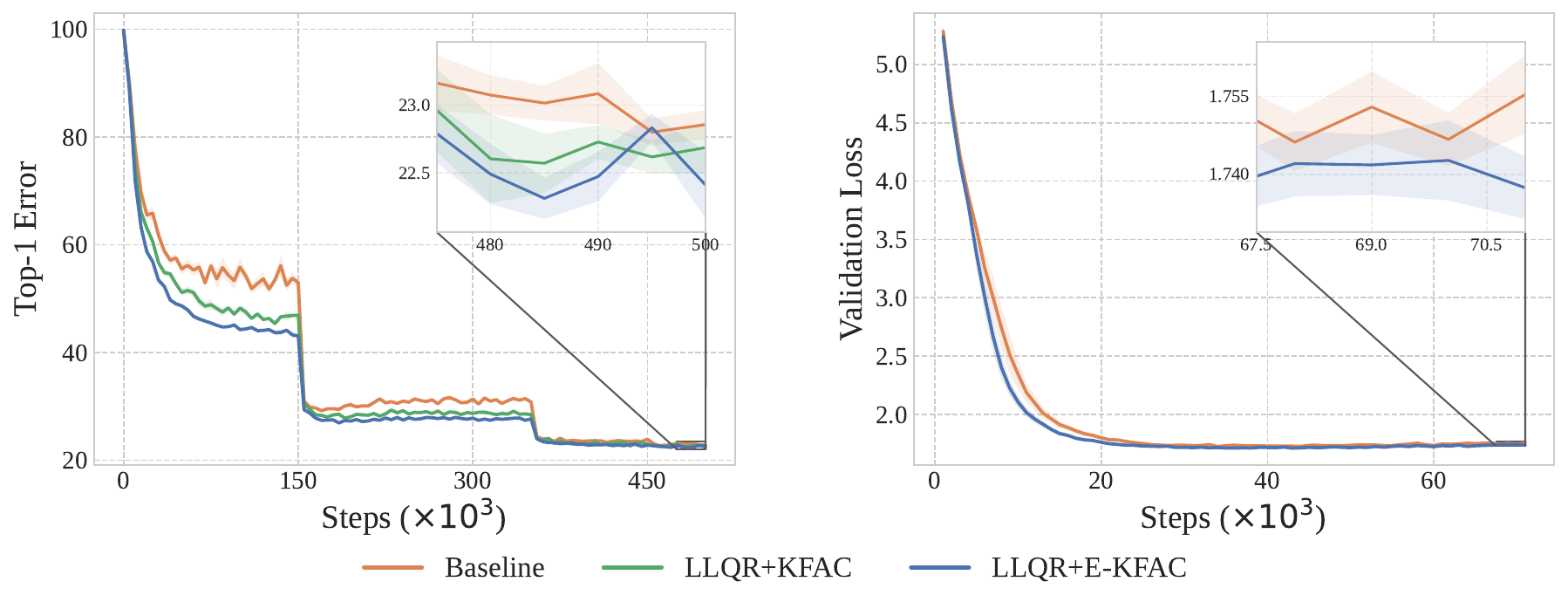}
  \caption{ImageNet and IWSLT14 training curves under NGD-induced divergence. \textbf{Left}: ResNet-50 trained for 100 epochs on ImageNet with SGDM, comparing the baseline, LLQR+KFAC, and LLQR+E-KFAC. \textbf{Right}: a Fairseq Transformer trained on IWSLT14 De--En with AdamW, comparing the baseline and LLQR+E-KFAC.} 
  \label{fig:imagenet_and_iwslt14_training_curves}
\end{figure*}

\section{Introduction}

Many successful optimization algorithms in deep learning can be understood as steepest-descent methods under a chosen geometry. For a differentiable loss $L(\vtheta)$ with gradient
$\vg = \nabla L(\vtheta)$, the gradient descent step is the solution of the variational problem
\begin{equation*}
\arg\min_{\Delta \vtheta}\left[ \vg^\top \Delta \vtheta + \frac{1}{2 \eta} \|\Delta \vtheta\|_2^2 \right] = - \eta \vg
\end{equation*}
corresponding to steepest descent in the Euclidean norm \citep{Boyd_Vandenberghe_2004}. Newton’s method uses the local loss Hessian; natural gradient uses the Fisher information matrix; and practical methods such as Adam, Shampoo, Muon, K-FAC, and related algorithms can be interpreted as using structured norms or preconditioners that capture partial information about the local geometry~\citep{DBLP:journals/corr/abs-2409-20325}. This viewpoint is increasingly important: the choice of geometry does not merely affect conditioning and convergence speed, but also changes the implicit bias of the training trajectory and can therefore influence the solutions found by learning~\citet{Pascanu2025PositionPaper}.


The classical difficulty is that the most principled geometries are dense and expensive. A generic second-order or geometry-aware step minimizes a local quadratic model
\begin{equation} \label{eq:steep_obj}
m(\Delta \vtheta) = \vg^\top \Delta \vtheta + \frac12 \Delta \vtheta^\top \mH \Delta \vtheta
\end{equation}
where $\mH$ may be the regularized Hessian, the Gauss–Newton matrix, the Fisher matrix, or another divergence-induced metric. 
For modern neural networks, $\mH$ is a dense matrix coupling parameters across layers through the chain rule. Directly solving $\mH \Delta \vtheta = -\vg$ is infeasible at scale. Existing scalable methods therefore introduce structure early:
for instance, 
K-FAC \citep{pmlr-v37-martens15} approximates the Fisher by block-diagonal Kronecker-factored blocks.  
This makes preconditioning tractable, but it also removes inter-layer couplings before the optimization problem defining the update has been solved.

This paper takes a different route. Rather than imposing structure directly on the curvature model, we first rewrite the dense geometry-aware update as a layer-coupled optimal-control problem.  
Building on the classical link between Newton steps and Linear Quadratic Regulator (LQR)~\citep{Dunn1989EfficientDP}, we show that under a broad class of divergence-induced quadratic models, the dense global quadratic form (\ref{eq:steep_obj}) admits an exact layerwise factorization: the forward pass defines linear perturbation dynamics, and the chosen divergence defines the LQR costs. 
The resulting problem retains the cross-layer dependencies of the original dense quadratic model, but expresses them through layerwise dynamics and costs rather than through an explicitly formed curvature matrix. 

The exact LQR formulation can be solved by Riccati recursions and recovers the corresponding exact geometry-aware update. However, for large neural networks, it still requires manipulating large Jacobian-dependent quantities and solving expensive matrix recursions. Our main contribution is therefore a scalable relaxation of this LQR problem. We parameterize the update as a preconditioned gradient,
\beq
\Delta\vtheta_i \;=\; \crtext{-}\mU_i\,\nabla_{\vtheta_i} L(\vtheta^k),
\eeq
where $\mU=\mathrm{diag}(\mU_0,\dots,\mU_{N-1})$ is a structured inverse preconditioner, for example diagonal or Kronecker-factored.
Crucially, this block structure is imposed on the learned inverse preconditioner, not on the geometry before deriving the update objective. This distinction is important. 
Unlike methods that begin by replacing the curvature matrix with a block-diagonal surrogate, our approach first derives a layer-coupled objective equivalent to the dense geometry-aware step, and only then restricts the class of inverse preconditioners used to approximate the resulting update. The learned blocks are therefore optimized in the presence of cross-layer couplings encoded by the LQR dynamics and cost matrices. This provides a systematic way to trade expressivity for scalability while retaining a principled connection to the exact second-order geometry.


Our {\bf contributions} are:

\begin{enumerate}
    \item \textbf{A layerwise optimal-control formulation of geometry-aware descent.}
    We show that steepest descent under a broad class of divergence-induced quadratic models—including Newton, Gauss--Newton, natural gradient, and intermediate-layer metrics—can be written exactly as a finite-horizon LQR problem. This reformulation separates the network dynamics from the choice of descent geometry and provides an exact reference update via Riccati recursions.


    \item \textbf{A scalable relaxation based on learned structured inverse preconditioners.}
    We propose to learn a structured inverse preconditioner directly by minimizing the LQR objective. This yields a practical family of geometry-aware optimizers that can use diagonal, Kronecker-factored, or other structured blocks, while preserving the layer-coupled objective induced by the original dense quadratic model.
    \item \textbf{A \bluetext{new} optimizer wrapper for modern architectures.} The relaxed LLQR update can be wrapped around standard optimizers such as \crtext{SGDM, AdamW, or Muon}, reusing the learned preconditioner across iterations and avoiding explicit curvature inversion. This makes the method flexible with respect to both the underlying divergence and the chosen preconditioner structure. 
    \item \textbf{Empirical validation on ResNets and Transformers.}
    Experiments show that LLQR improves convergence and final performance on image classification \bluetext{and translation benchmarks,} accelerates grokking in Transformers, \crtext{and complements AdamW and Muon in autoregressive language modeling,} while adding only modest computational overhead. The results position LLQR as a practical and extensible framework for studying geometry-aware optimization at scale.

\end{enumerate}

\section{Related Work}
The connection between deep learning and optimal control has been recognized since the early days, with back-propagation framed as a control problem and linked to automatic differentiation \citep{bryson1969applied, athans2013optimal, Cun1988ATF}.
More recently, neural networks have been cast as discrete-time nonlinear dynamical systems, where layers correspond to time steps and weights to control variables \citep{Weinan2017APO}.
Within this view, tools such as Pontryagin’s Maximum Principle and Differential Dynamic Programming have been used to derive new training algorithms \citep{DBLP:journals/jmlr/LiCTE17, DBLP:conf/icml/LiH18, DBLP:conf/iclr/LiuCT21}.
Our work differs in focus: rather than reformulating the training problem itself, we recast the steepest-descent step as an optimal control problem.

Closer to our approach, \citet{DBLP:conf/amcc/MizutaniD05} proposed a stagewise Newton method for multilayer perceptrons, relying on classical results from \citet{dreyfus1966numerical}---equivalent to \citet{Dunn1989EfficientDP} that we extend---to propagate Hessian information through the state space without explicitly forming Newton updates.
Extensions followed \citep{DBLP:conf/ijcnn/MizutaniDD05, DBLP:conf/ijcnn/MizutaniD06, DBLP:journals/nn/MizutaniD08}, but these remained confined to Newton’s method and were not generalized to other geometries.
In contrast, our formulation covers a broad family of quadratic models, including Gauss–Newton, Fisher-based (natural gradient), and intermediate-layer metrics.

Natural gradient descent (NGD) has been especially influential in deep learning \citep{Amari1998NaturalGW, Pascanu2014RevisitingNG}, though its practical use is limited by the cost of computing and inverting the full Fisher information matrix.
K-FAC \citep{pmlr-v37-martens15} addressed this by approximating the Fisher with block-diagonal Kronecker-factored surrogates, enabling scalable preconditioning.
Follow-up work improved efficiency and robustness through refined factorization and inversion schemes \citep{EKFAC--GeorgeLBBV18, DBLP:conf/mlsys/Osawa0H23, DBLP:conf/icml/LinDENKTM24, GomesZBWH25adafisher}.
These methods approximate the preconditioner by imposing structure on the curvature matrix; our approach instead frames the inverse preconditioner itself as an optimization variable---with block structure imposed on the parametrization---learned under the LQR formulation.

\section{Problem Formulation}

We consider a minibatch of size $B$. Throughout the paper, $\vx_i$
denotes the full batched activation at layer $i$, i.e.
$\vx_i \in \mathbb{R}^{B \times d_i}$, with rows
$\vx_i^{(b)}$ corresponding to individual samples. For a feedforward network of depth $N$, the network dynamics are
\[
\vx_{i+1}^{(b)} = f_i(\vx_i^{(b)},\theta_i),
\qquad i=0,\dots,N-1,
\]
or, equivalently, in batched notation,
\beq \label{eq:forward_pass}
\vx_{i+1}=f_i(\vx_i,\vtheta_i),
\eeq
where the same parameter $\vtheta_i$ is shared across all samples. At iteration $k$, the training loss is the empirical minibatch loss $L(\vtheta^k) \equiv \ell(\vx_N(\vtheta^k))$
where $\ell$ acts primarily on the outputs $\vx_N$.\footnote{
The formulation extends directly when $\ell$ also depends on intermediate activations $\vx_i(\vtheta)$ and includes explicit regularization terms in $\vtheta_i$.} Thus, all gradients and quadratic models below are defined with respect to
the minibatch objective.


At iterate $\vtheta^k$, we consider the steepest-descent step
\begin{equation} \label{eq:quadratic_model}
\arg\min_{\Delta\vtheta} \nabla L(\vtheta^k)^\top \Delta \vtheta \;+\;
\tfrac12 \Delta \vtheta^\top \mH(\vtheta^k) \Delta \vtheta ,
\end{equation}
where $\mH$ is a symmetric matrix-valued function of the parameters.
Whenever $\mH$ is positive definite (or suitably regularized), the solution is
$\Delta\vtheta^{\star} = -\mH^{-1} \vg$.

\paragraph{Divergence-induced quadratic models.}
Rather than arbitrary quadratic forms, we restrict to those arising as second-order expansions of divergences with layerwise structure.

\begin{definition}[Divergence-Induced Quadratic Models]
\label{def:diq}
Fix the iterate $\vtheta^k$ and its (batched) activations
$\{\vx_i(\vtheta^k)\}_{i=0}^N$.
A quadratic model is \emph{divergence-induced} if it is the Hessian (in $\vtheta$) of a divergence of the form:
\begin{equation}
\label{eq:layerwise}
D(\vtheta^k,\vtheta)
\;=\;
\psi_N\!\left(\vx_N(\vtheta)\right)
\;+\;
\sum_{i=0}^{N-1}
\psi_i\!\left(\vx_i(\vtheta),\, \vtheta_i\right),
\end{equation}
where each $\psi_i,\psi_N$ is twice continuously differentiable.\footnote{Dependence on the basepoint $\vtheta^k$ (and thus the minibatch at iteration $k$) is implicit and omitted for clarity.}
This form ensures that the dependence on $\vtheta$ is either direct
(layerwise through $\vtheta_i$) or mediated via forward activations $\vx_i(\vtheta)$.
The associated local quadratic model is:
\begin{equation} \label{eq:div_hessian}
\mH := \nabla^2_{\vtheta\vtheta} D(\vtheta^k,\vtheta)\big|_{\vtheta=\vtheta^k}
\end{equation}
\end{definition}
This family covers nearly all geometry-aware methods of interest, including Newton, Gauss--Newton, Fisher, and their regularized variants (see Appendix \ref{app:steepest-descent-as-lqr-detailed}).

\paragraph{Goal.}
Compute the update (\ref{eq:quadratic_model}) exactly for this class of models, without forming or inverting the dense global $\mH$, by exploiting the sequential structure of the network.

\section{Steepest Descent as LLQR}\label{sec:steepest-descent-as-llqr}
In this section, we follow the classical optimal-control derivation for Newton steps \citep[e.g.,][Ch.~2.6]{bertsekas2016nonlinear},
extending it to the much broader family of divergence-induced quadratic models.
Full details are given in Appendix~\ref{app:steepest-descent-as-lqr-detailed}.

\subsection{Layerwise linearization}
\label{sec:linearization}

\paragraph{Notation}
At the iterate $\vtheta_i^k$, define parameter perturbations
$
\delta\vtheta_i := \vtheta_i - \vtheta_i^k
$
so that the global update is
$\Delta \vtheta = \mathrm{concat}(\delta \vtheta_0,\dots,\delta \vtheta_{N-1})$.
We also introduce \emph{linearized activation perturbations} $\delta\vx_i$, obtained by expanding each layer map $f_i$ at $(\vx_i^k,\vtheta_i^k)$:
\begin{equation}\label{eq:linearization}
\delta \vx_{i+1} \;=\; \mA_i \,\delta \vx_i + \mB_i \,\delta \vtheta_i,\quad \delta \vx_0 = 0, \qquad
\mA_i := \frac{\partial f_i}{\partial \vx_i}(\vx_i^k,\vtheta_i^k), \quad \mB_i := \frac{\partial f_i}{\partial \vtheta_i}(\vx_i^k,\vtheta_i^k).
\end{equation}
In the batched setting, $\delta \vx_i \in \mathbb{R}^{B \times d_i}$ is
batch-indexed, whereas $\delta\vtheta_i$ is shared across the batch; $\mA_i$ is block-diagonal over samples and $\mB_i$ stacks the samplewise parameter Jacobians.

We note (see Appendix~\ref{sec:gd-as-lqr}) that by the chain rule, the first-order variations
\begin{equation}
\label{eq:xvariations}
\delta \vx_i:=\nabla_{\vtheta} \vx_i^\top \Delta \vtheta
\end{equation}
are solutions of the recursion (\ref{eq:linearization}) when optimizing \eqref{eq:quadratic_model} with $\mH(\vtheta^k)=\mI$.

\subsection{From global quadratics to layerwise LQR}\label{sec:layerwise-lqr-finite-horizon}

We now express the linear and quadratic terms of \eqref{eq:quadratic_model} in terms of $(\delta\vx_i,\delta\vtheta_i)$ and the dynamics \eqref{eq:linearization}.
This yields a quadratic program with per-layer cost and linear dynamics. 
For notational clarity, we present the derivation for a single input ($B=1$). The minibatch case is obtained by treating $\vx_i$ and $\delta\vx_i$ as stacked batch activations and perturbations---while $\delta\vtheta_i$ remains shared across all examples---and solving with respect to the whole minibatch. Accordingly, $\mA_i$ and $\mB_i$ are the Jacobians of the batched layer map, and $\mQ_i,\mM_i,\mR_i$ are the corresponding Hessian blocks of the minibatch objective with respect to $(\delta\vx_i,\delta\vtheta_i)$.

\paragraph{Linear term.}
By the chain rule and (\ref{eq:xvariations}),
\begin{equation}\label{eq:linear_term}
\vg^\top\Delta\vtheta \;=\; \nabla \ell(\vx^k_N)^\top \delta\vx_N
\end{equation}

\paragraph{Quadratic term.}
To make the Hessian (\ref{eq:div_hessian}) decomposable by layer, we augment the divergence with multipliers $\vp = [p_1, \cdots p_{N}]$ for the constraints \eqref{eq:linearization}, defining
\begin{equation}\label{eq:augmented-divergence}
\mathcal{H}(\vx,\vtheta,\vp) = \psi_N(\vx_N) + \sum_{i=0}^{N-1} \Big[\psi_i(\vx_i,\vtheta_i) + \vp_{i+1}^\top\!\big(f_i(\vx_i,\vtheta_i)-\vx_{i+1}\big)\Big].
\end{equation}
For any choice of $\vp$, $D(\vtheta^k,\vtheta)=\mathcal{H}(\vx(\vtheta),\vtheta,\vp)$.
Choosing $\vp$ to enforce stationarity of $\cH$ in the activation variables cancels cross-derivative terms and simplifies the quadratic expansion:

\begin{theorem} \label{theo:LLQR}
Let $\vp^k$ satisfy the adjoint recursion
\[
\vp^k_N = \nabla_{\vx_N} \psi_N(\vx_N^k), \qquad \vp^k_i = \mA_i^\top \vp^k_{i+1} + \nabla_{\vx_i} \psi_i(\vx_i^k,\vtheta_i^k).
\]
Then $\nabla_{\vx} \cH\big(\vx(\vtheta^k),\vtheta^k, \vp^k\big) = 0$, and the quadratic term in (\ref{eq:quadratic_model}) decomposes layerwise:
\begin{equation*}
\Delta \vtheta^\top \mH \Delta \vtheta
= \delta\vx_N^\top \mQ_N \delta\vx_N
+ \sum_{i=0}^{N-1}
\begin{bmatrix}\delta\vx_i \\ \delta\vtheta_i\end{bmatrix}^{\!\top}
\begin{bmatrix}\mQ_i & \mM_i^\top \\ \mM_i & \mR_i\end{bmatrix}
\begin{bmatrix}\delta\vx_i\\ \delta\vtheta_i\end{bmatrix}
\end{equation*}
subject to the constraints (\ref{eq:linearization}), with local blocks below evaluated at $\vx^k, \vtheta^k, \vp^k$.
\[
\mQ_i = \nabla^2_{\vx_i\vx_i} \cH,\qquad \mM_i = \nabla^2_{\vtheta_i\vx_i} \cH,\qquad \mR_i = \nabla^2_{\vtheta_i\vtheta_i} \cH,
\]
\end{theorem}

Theorem~\ref{theo:LLQR} generalizes classical results from \citet{Dunn1989EfficientDP, PantojaM1989}, extending the Newton–LQR connection to divergence-induced steepest descent. The proof is given in Appendix~\ref{app:steepest-descent-as-lqr-detailed}.

\paragraph{Sequential quadratic program}
Combining (\ref{eq:linear_term}) and Theorem~\ref{theo:LLQR}, the steepest-gradient step~\eqref{eq:quadratic_model} is thus equivalent to 
\begin{equation}
\label{eq:lqr-form}
\begin{aligned}
\min_{\delta\vtheta_i}\quad
& \nabla_{\vx_N}\ell(\vx^k_N)^\top
 \delta\vx_N
  + \tfrac12\,\delta\vx_N^\top \mQ_N \delta\vx_N + \tfrac12\sum_{i=0}^{N-1}
\begin{bmatrix}\delta\vx_i \\ \delta\vtheta_i\end{bmatrix}^\top
\begin{bmatrix}\mQ_i &  \mM_i^\top \\ \mM_i & \mR_i\end{bmatrix}
\begin{bmatrix}\delta\vx_i\\ \delta\vtheta_i\end{bmatrix} 
\\[6pt]
\text{s.t.}\quad
& \delta\vx_{i+1} = \mA_i\delta\vx_i + \mB_i\delta\vtheta_i,
   \qquad \delta\vx_0=\mathbf{0}.
\end{aligned}
\end{equation}
This is precisely a finite-horizon Linear Quadratic Regulator. 

\subsection{KKT conditions and Riccati recursions}
The LQR problem (\ref{eq:lqr-form}) 
can be solved efficiently by Riccati recursion \citep{kalman1960contributions, reid1972riccati}.
Introducing new multipliers $\vq := [\vq_1, \cdots, \vq_N]$ for the linear constraints, stationarity of the resulting Lagrangian with respect to $\delta \vtheta_i$ and $\delta \vx_i$ gives respectively, with 
$\vq_N = \mQ_N \delta \vx_N + \vg_N$:
\begin{align}
\mR_i \delta \vtheta_i + \mM_i \delta \vx_i + \mB_i^\top \vq_{i+1} &= 0 \label{eq:KKT-dtheta} \\
\mQ_i \delta \vx_i + \mM_i^\top \delta \vtheta_i + \mA_i^\top \vq_{i+1} - \vq_i &= 0
\label{eq:KKT-dx}
\end{align}

To solve this, posit the quadratic form $
\vq_i = \mK_i \delta \vx_i + \vlambda_i,
$
where $\mK_i$ and $\vlambda_i$ are to be determined.
Substituting into \eqref{eq:KKT-dtheta} and \eqref{eq:KKT-dx} yields backwards recursions for $\mK_i, \vlambda_i$ given in Appendix~\ref{secapp:riccati}. 
%
The optimal updates are 
then obtained 
by a forward rollout
using \eqref{eq:linearization}.

The Riccati solution illustrates how the layerwise structure of the network can be exploited directly:
a backward pass computes local gain matrices ($\mK_i$) and adjoints ($\vlambda_i$),
and a forward pass assembles the update.
This replaces the cubic $\mathcal{O}(d^3)$ cost of dense inversion by a sequence of smaller layerwise solves, scaling as $\mathcal{O}(\sum_i d_i^3)$ when $\sum_i d_i=d$. We view this exact solution primarily as a reference for the geometry-aware update defined by the original quadratic model, and validate the equivalence on a toy problem in Section~\ref{sec:Riccati_toyexp}. 

For modern networks, however, running Riccati recursions remains impractical. The exact formulation is therefore most useful as a derivation device: it identifies the layerwise dynamics and cost matrices that encode the dense geometry, and motivates our main algorithmic contribution below, 
a scalable relaxation that learns a structured inverse preconditioner under the resulting LQR objective.

\section{Relaxation}\label{sec:relaxation-and-algo}



Instead of computing $\delta\vtheta_i$ exactly from the Riccati backward pass, we thus replace it by a preconditioned gradient with a learned inverse preconditioner $\mU_i := \mP_i^{-1}$, i.e., 
$
\delta\vtheta_i \;=\; -\mU_i\,\nabla_{\vtheta_i} L(\vtheta^k)
$,
with block-diagonal $\mU=\mathrm{diag}(\mU_0,\dots,\mU_{N-1})$.
Substituting into the LQR objective yields the relaxed problem
\begin{equation*}
\min_{\mU} \quad
 \nabla_{\vx_N}\ell(\vx_N)^\top \delta\vx_N
+ \tfrac12\,\delta\vx_N^\top \mQ_N \delta\vx_N 
\,+\sum_{i=0}^{N-1}\!
\Big(\tfrac12\,\delta\vx_i^\top \mQ_i \delta\vx_i
+ \tfrac12\,\delta\vtheta_i^\top \mR_i \delta\vtheta_i
+ \delta\vtheta_i^\top \mM_i \delta\vx_i\Big)
\end{equation*}
\vspace{-0.2cm}
\begin{equation}
\label{eq:relaxed-lqr-equivalence}
\text{s.t.}\quad
 \delta\vx_{i+1} = \mA_i \delta\vx_i - \mB_i \mU_i\nabla_{\vtheta_i} L(\vtheta^k),
\: \delta\vx_0=\mathbf{0}.
\end{equation}

\paragraph{Advantages.}
Optimizing directly over $\mU$ has three benefits:
(i) It provides the \emph{inverse} preconditioner directly, avoiding costly matrix inversion.
(ii) It can be recomputed every $n$ iterations, requiring only cheap layerwise multiplications with gradients in between.
(iii) Structure (e.g.\ diagonal, Kronecker) can be imposed on $\mU_i$ to further reduce cost and memory.

\paragraph{Remark.}
Problem \eqref{eq:relaxed-lqr-equivalence} admits many equivalent solutions, since $\mU$ is only constrained along directions spanned by the gradient. This mismatch across iterations creates a gap between the exact LLQR and the relaxed update.
To mitigate it, Algorithm~\ref{alg:relaxed-lqr-sd} (i) restricts $\mU$ structurally, (ii) optimizes it over the full minibatch with $\vtheta^k$ fixed, and (iii) \mdfy{uses Exponential Moving Average (EMA) to smoothly update from the previous solution.}

In Appendix~\ref{app:algo-details}, we explain how Algorithm~\ref{alg:relaxed-lqr-sd} can be efficiently implemented using automatic differentiation \citep{DBLP:journals/neco/Pearlmutter94, DBLP:journals/jmlr/BaydinPRS17}. \bluetext{In particular, we expose in Appendix~\ref{app:chunking} how a chunking strategy over Eq.~\eqref{eq:relaxed-lqr-equivalence} can be incorporated to efficiently trade memory for compute when needed.} \mdfy{We also present the ablations performed to tune the method,} and detail how damping can be incorporated.
\begin{algorithm}
  \caption{LLQR: Layerwise LQR Steepest Descent (relaxed periodic preconditioner)}
  \label{alg:relaxed-lqr-sd}
\scriptsize
\begin{algorithmic}[1]
\REQUIRE model $f$, loss $L$; data loader $\mathcal{D}$; outer optimizer $\mathcal{O}_{\mathrm{out}}$; step size $\eta$; layer index $i$
\REQUIRE recompute period $n$; inner iters $T$; structure $\mathcal{S}$ (e.g., diag/Kronecker); inner optimizer $\mathcal{O}_{\mathrm{in}}$; inner step size $\alpha$; EMA parameter $\beta$ 
\STATE Initialize $\vtheta^0$, set $\mU\leftarrow \mI$, project to $\mathcal{S}$
\FOR{$k = 0,1,2,\dots$}
  \STATE Sample minibatch $(\vx^k,\vy^k)\sim\mathcal{D}$ and compute $g_k \leftarrow \nabla_{\vtheta} L(\vtheta^k)$
  \IF{$k \bmod n = 0$}
     \STATE Linearize $(\mA_i,\mB_i) \gets (\nabla_{\vx}, \nabla_{\vtheta}) f_i(\vx_i^k,\vtheta_i^k)$ \ALGCOMMENT{(Re)fit preconditioner at $\vtheta^k$}
     \STATE Form blocks $(\mQ_i,\mR_i,\mM_i)$
     \STATE $\mU_0 \gets \mI$
     \FOR{$t = 0$ to $T-1$}
        \STATE $\mU_{t+1} \gets \mathcal{O}_{\mathrm{in}}\!\big(\mU_t, \nabla_{\mU_t} J, \alpha\big)$ \ALGCOMMENT{$J$: relaxed obj.\ \eqref{eq:relaxed-lqr-equivalence}}
     \ENDFOR
     \STATE $\mU \gets \beta\mU + (1-\beta)\mU_T$ \ALGCOMMENT{EMA update}
  \ENDIF
  \STATE Preconditioned step: $\Delta\vtheta_k \gets -\mU g_k$
  \STATE Update params: $\vtheta^{k+1} \gets \mathcal{O}_{\mathrm{out}}(\vtheta^k,\Delta\vtheta_k,\eta)$
\ENDFOR
\end{algorithmic}
\end{algorithm}

\section{Empirical Validation}

We now evaluate LLQR across several settings to demonstrate its flexibility as a framework for geometry-aware preconditioning. We vary both the structure of the learned inverse preconditioner $\mU$ and the divergence defining the LQR objective. Specifically, we consider Kronecker-factored, E-KFAC, and diagonal preconditioners (Appendix~\ref{app:block_structures}), together with two geometries: the KL divergence, which yields an approximate natural-gradient update, and the Bregman gap of the regularized loss, which yields an approximate Newton update (Appendix~\ref{app:damped_Newton}). \crtext{We also evaluate LLQR with AdamW and Muon in autoregressive language modeling (Section~\ref{sec:autoregressive_lm}), and examine objective coupling and hyperparameter sensitivity in Appendices~\ref{app:coupled-objective-ablation} and~\ref{app:sensitivity_sweeps}, respectively.}



\subsection{Exact LLQR: a toy example}
\label{sec:Riccati_toyexp}
We first use Rosenbrock's function---a classic non-convex benchmark with a narrow curved valley that is challenging for gradient descent---to validate both the exact LQR equivalence and the relaxed LLQR approximation. Reformulating it as a two-layer network (Appendix~\ref{app:rosenbrock_details}), we compare Newton's step with the update obtained by solving \eqref{eq:lqr-form} through the Riccati recursions 
; the resulting trajectories match exactly, confirming the equivalence. We then apply the relaxation to learn a block-diagonal inverse preconditioner $\mU$ with one scalar per layer, observing that more frequent preconditioner updates accelerate convergence and make the trajectory increasingly Newton-like.
Returning to the single-layer Rosenbrock formulation, we further use a dense structure for $\mU$ and recover a convergence rate matching the exact Newton step. Results are shown in Fig.~\ref{fig:rosenbrock} of Appendix~\ref{app:rosenbrock_details}. 

\bluetext{To illustrate that LLQR can encode inter-layer couplings even under restricted structure, Fig.~\ref{fig:rosenbrock_diagnostics} in Appendix~\ref{app:further_exps} compares the two-layer LLQR trajectory with Newton's method, exact LQR, and an approximate Newton update based only on the diagonal Hessian. We observe that LLQR more closely tracks the Newton and exact-LQR trajectories, while the diagonal-Hessian update fails to converge and becomes trapped where the full Rosenbrock Hessian is indefinite. Both methods use diagonal degrees of freedom, but LLQR learns them through an update-aligned curvature correction: by optimizing toward the induced step direction, it preserves the effect of inter-layer couplings that $\operatorname{diag}(\mH)$ discards before inversion. 
}

\begin{table}[tb]
\centering
\caption{Performance and wall-clock time ratio on ResNet-18.
\textcolor{BrickRed}{Newton} and \textcolor{RoyalBlue}{NGD} denote the divergence-induced quadratic used by LLQR. Across our experiments, $\pm$ denotes the standard error.}
\label{tab:cifar-performance}
\setlength{\tabcolsep}{3.0pt}
\renewcommand{\arraystretch}{0.92}
\scriptsize
\resizebox{\linewidth}{!}{%
\begin{tabular}{lllccccccc}
\toprule
\textbf{Dataset} & \textbf{Optimizer} & \textbf{Metric}
& \textbf{Base}
& \multicolumn{2}{c}{\textbf{Diagonal}}
& \multicolumn{2}{c}{\textbf{K-FAC}}
& \multicolumn{2}{c}{\textbf{E-KFAC}}\\
\cmidrule(lr){5-6}\cmidrule(lr){7-8}\cmidrule(lr){9-10}
& & & BS = 128
& \textcolor{BrickRed}{Newton}
& \textcolor{RoyalBlue}{NGD}
& \textcolor{BrickRed}{Newton}
& \textcolor{RoyalBlue}{NGD}
& \textcolor{BrickRed}{Newton}
& \textcolor{RoyalBlue}{NGD}\\
\midrule
\textbf{CIFAR--10} & SGDM & Top-1 Acc.
& $96.19{\scriptscriptstyle \,\pm\, 0.07}$
& $96.25{\scriptscriptstyle \,\pm\, 0.04}$
& $96.29{\scriptscriptstyle \,\pm\, 0.07}$
& $96.29{\scriptscriptstyle \,\pm\, 0.10}$
& $96.33{\scriptscriptstyle \,\pm\, 0.19}$
& $\mathbf{96.37}{\scriptscriptstyle \,\pm\, 0.05}$
& $\mathbf{96.37}{\scriptscriptstyle \,\pm\, 0.05}$\\
& & Time
& $\times 1.0$
& $\times 1.08$
& $\times 1.08$
& $\times 1.22$
& $\times 1.22$
& $\times 1.34$
& $\times 1.31$\\

& AdamW & Top-1 Acc.
& $94.54{\scriptscriptstyle \,\pm\, 0.11}$
& $94.64{\scriptscriptstyle \,\pm\, 0.09}$
& $94.55{\scriptscriptstyle \,\pm\, 0.06}$
& $94.73{\scriptscriptstyle \,\pm\, 0.10}$
& $94.67{\scriptscriptstyle \,\pm\, 0.18}$
& $94.72{\scriptscriptstyle \,\pm\, 0.04}$
& $\mathbf{94.95}{\scriptscriptstyle \,\pm\, 0.05}$\\
& & Time
& $\times 1.0$
& $\times 1.01$
& $\times 1.01$
& $\times 1.05$
& $\times 1.04$
& $\times 1.12$
& $\times 1.15$\\
\midrule
\textbf{CIFAR--100} & SGDM & Top-1 Acc.
& $78.97{\scriptscriptstyle \,\pm\, 0.22}$
& $79.21{\scriptscriptstyle \,\pm\, 0.09}$
& $79.20{\scriptscriptstyle \,\pm\, 0.30}$
& $79.42{\scriptscriptstyle \,\pm\, 0.25}$
& $79.22{\scriptscriptstyle \,\pm\, 0.11}$
& $79.83{\scriptscriptstyle \,\pm\, 0.08}$
& $\mathbf{79.96}{\scriptscriptstyle \,\pm\, 0.10}$\\
& & Time
& $\times 1.0$
& $\times 1.08$
& $\times 1.08$
& $\times 1.22$
& $\times 1.23$
& $\times 1.31$
& $\times 1.32$\\

& AdamW & Top-1 Acc.
& $76.09{\scriptscriptstyle \,\pm\, 0.27}$
& $76.02{\scriptscriptstyle \,\pm\, 0.31}$
& $76.28{\scriptscriptstyle \,\pm\, 0.13}$
& $76.32{\scriptscriptstyle \,\pm\, 0.39}$
& $76.49{\scriptscriptstyle \,\pm\, 0.06}$
& $76.96{\scriptscriptstyle \,\pm\, 0.03}$
& $\mathbf{77.07}{\scriptscriptstyle \,\pm\, 0.16}$\\
& & Time
& $\times 1.0$
& $\times 1.01$
& $\times 1.01$
& $\times 1.05$
& $\times 1.06$
& $\times 1.13$
& $\times 1.13$\\
\bottomrule
\end{tabular}
}
\vspace{-0.5em}
\end{table}

\subsection{Experiments on CIFAR--10/100}

We evaluate the relaxed LLQR formulation (Algorithm~\ref{alg:relaxed-lqr-sd}) on modern architectures by training ResNet-18~\citep{he2016deep} on CIFAR--10 and CIFAR--100~\citep{krizhevsky2009learning}. LLQR is used as a wrapper around SGD with momentum (SGDM) and AdamW, and compared against carefully tuned versions of each base optimizer (Appendix~\ref{app:CIFAR_experiments_details}). When adding LLQR, we keep the base-optimizer hyperparameters unchanged; the method therefore acts only through the learned gradient preconditioner.


Although this setup favors the unmodified baselines, LLQR improves iteration-wise convergence when using Kronecker or E-KFAC preconditioner structures (Fig.~\ref{fig:cifar10_cifar100_top1_error_curves}) and remains competitive in wall-clock time, especially during early and mid-training. In contrast, the diagonal structure provides little noticeable acceleration, suggesting that it may be insufficiently expressive to capture the relevant layerwise curvature. Overall, LLQR adds only a modest training-time overhead while yielding small but consistent final-performance gains (Table~\ref{tab:cifar-performance}).
\crtext{To investigate further the inter-layer couplings outside of toy settings, we compare the full objective with its block-diagonal restriction (Appendix~\ref{app:coupled-objective-ablation}) under a matched ResNet-18 fitting schedule, showing the benefits of keeping cross-block curvature within the fitting objective.}

\bluetext{We also extend the CIFAR evaluation to PyramidNet-110, VGG-16-BN, and WideResNet-28-10, all using the E-KFAC block structure. For PyramidNet-110, chunking is necessary (Appendix~\ref{app:chunking}) to keep the inner-loop batch size equal to the outer-loop batch size, causing a small slowdown but validating the chunking strategy. We retain the ResNet hyperparameters throughout, sweeping only weight decay for the baselines. LLQR hyperparameters are kept fixed as in the ResNet experiments, demonstrating robust transfer across architectures (Table~\ref{tab:cifar-architecture-performance}).}

\begin{table}[tb]
\centering
\caption{Performance and wall-clock time ratio across CIFAR architectures with E-KFAC block structure.
\textcolor{BrickRed}{Newton} and \textcolor{RoyalBlue}{NGD} denote the divergence-induced quadratic used by LLQR.}
\label{tab:cifar-architecture-performance}
\setlength{\tabcolsep}{3.0pt}
\renewcommand{\arraystretch}{0.92}
\scriptsize
\resizebox{\linewidth}{!}{%
\begin{tabular}{lllccccccccc}
\toprule
\textbf{Dataset} & \textbf{Optimizer} & \textbf{Metric}
& \multicolumn{3}{c}{\textbf{PyramidNet-110}}
& \multicolumn{3}{c}{\textbf{VGG-16-BN}}
& \multicolumn{3}{c}{\textbf{WRN-28-10}}\\
\cmidrule(lr){4-6}\cmidrule(lr){7-9}\cmidrule(lr){10-12}
& & & \textbf{Base}
& \textcolor{BrickRed}{Newton}
& \textcolor{RoyalBlue}{NGD}
& \textbf{Base}
& \textcolor{BrickRed}{Newton}
& \textcolor{RoyalBlue}{NGD}
& \textbf{Base}
& \textcolor{BrickRed}{Newton}
& \textcolor{RoyalBlue}{NGD}\\
\midrule
\textbf{CIFAR--10} & SGDM & Top-1 Acc.
& $97.18{\scriptscriptstyle \,\pm\, 0.08}$
& $\mathbf{97.26}{\scriptscriptstyle \,\pm\, 0.04}$
& $97.21{\scriptscriptstyle \,\pm\, 0.04}$
& $95.10{\scriptscriptstyle \,\pm\, 0.06}$
& $\mathbf{95.41}{\scriptscriptstyle \,\pm\, 0.05}$
& $95.27{\scriptscriptstyle \,\pm\, 0.09}$
& $96.94{\scriptscriptstyle \,\pm\, 0.04}$
& $\mathbf{97.10}{\scriptscriptstyle \,\pm\, 0.01}$
& $97.04{\scriptscriptstyle \,\pm\, 0.04}$\\
& & Time
& $\times 1.0$
& $\times 1.41$
& $\times 1.41$
& $\times 1.0$
& $\times 1.18$
& $\times 1.17$
& $\times 1.0$
& $\times 1.26$
& $\times 1.26$\\
\midrule
\textbf{CIFAR--100} & SGDM & Top-1 Acc.
& $83.96{\scriptscriptstyle \,\pm\, 0.11}$
& $84.00{\scriptscriptstyle \,\pm\, 0.07}$
& $\mathbf{84.40}{\scriptscriptstyle \,\pm\, 0.21}$
& $75.65{\scriptscriptstyle \,\pm\, 0.12}$
& $76.31{\scriptscriptstyle \,\pm\, 0.05}$
& $\mathbf{76.32}{\scriptscriptstyle \,\pm\, 0.16}$
& $82.50{\scriptscriptstyle \,\pm\, 0.08}$
& $82.72{\scriptscriptstyle \,\pm\, 0.11}$
& $\mathbf{82.80}{\scriptscriptstyle \,\pm\, 0.17}$\\
& & Time
& $\times 1.0$
& $\times 1.41$
& $\times 1.41$
& $\times 1.0$
& $\times 1.19$
& $\times 1.17$
& $\times 1.0$
& $\times 1.25$
& $\times 1.25$\\
\bottomrule
\end{tabular}
}
\vspace{-0.5em}
\end{table}

\subsection{\mdfy{ResNet-50 on ImageNet}}
\begin{wraptable}{r}{0.50\linewidth}
\vspace{-3.0em}
\centering
\caption{Performance and wall-clock time ratio on ResNet-50/ImageNet trained for 100 epochs and Fairseq Transformer IWSLT14 German-to-English translation.}
\label{tab:imagenet_iwslt14_results}
\setlength{\tabcolsep}{3.0pt}
\renewcommand{\arraystretch}{0.92}
\scriptsize
\resizebox{\linewidth}{!}{%
\begin{tabular}{lllcccc}
\toprule
\textbf{Benchmark} & \textbf{Metric} & \textbf{Optimizer}
& \textbf{Block} & \textbf{Div.}
& \textbf{Result} & \textbf{Time}\\
\midrule
\textbf{ImageNet} & Top-1 Acc. & SGDM
& --
& --
& $77.42{\scriptscriptstyle \,\pm\, 0.12}$
& $\times 1.00$\\
(100 epochs) & & SGDM
& K-FAC
& \textcolor{BrickRed}{Newton}
& $77.78{\scriptscriptstyle \,\pm\, 0.31}$
& $\times 1.027$\\
& & SGDM
& E-KFAC
& \textcolor{BrickRed}{Newton}
& $76.93{\scriptscriptstyle \,\pm\, 0.95}$
& $\times 1.032$\\
& & SGDM
& K-FAC
& \textcolor{RoyalBlue}{NGD}
& $77.98{\scriptscriptstyle \,\pm\, 0.08}$
& $\times 1.027$\\
& & SGDM
& E-KFAC
& \textcolor{RoyalBlue}{NGD}
& $\mathbf{78.05}{\scriptscriptstyle \,\pm\, 0.05}$
& $\times 1.032$\\
\midrule
\textbf{IWSLT14 De--En} & BLEU & AdamW
& --
& --
& $34.24{\scriptscriptstyle \,\pm\, 0.12}$
& $\times 1.00$\\
& & AdamW
& E-KFAC
& \textcolor{RoyalBlue}{NGD}
& $\mathbf{34.51}{\scriptscriptstyle \,\pm\, 0.05}$
& $\times 1.16$\\
\bottomrule
\end{tabular}
}
\vspace{-1.2em}
\end{wraptable}

\bluetext{
We evaluate LLQR on ImageNet \citep{imagenetDeng2009} by training a ResNet-50 for 100 epochs (Appendix~\ref{app:imagenet_details}). As shown in Fig.~\ref{fig:imagenet_and_iwslt14_training_curves} (left) and Table~\ref{tab:imagenet_iwslt14_results}, \textbf{LLQR reaches $\mathbf{78.05\pm0.05}$} top-1 accuracy, compared to $77.42\pm0.12$ for the SGDM baseline, while adding only a modest computational overhead of $\approx1.03\times$ per epoch on ImageNet (see Appendix~\ref{app:compute_overhead}). This overhead is smaller than on CIFAR datasets because the preconditioner is updated at the same per-epoch frequency while ImageNet epochs are substantially longer. When scaling from CIFAR--100 to ImageNet, the same hyperparameter choices worked best, suggesting that LLQR's hyperparameters are fairly robust across dataset scales.
}

\begin{table}[tb]
\centering
\caption[Practical comparison on CIFAR--100 and ImageNet]{Performance and wall-clock time comparison on ResNet-18/CIFAR--100 and ResNet-50/ImageNet, with batch size set to \textbf{256} across experiments to match benchmarks. \textcolor{RoyalBlue}{LLQR} denotes an E-KFAC block structure with NGD updates. Results marked with \textdagger follow \citet{GomesZBWH25adafisher} and enable efficiency comparisons with other approximate second-order methods. We report \textbf{time per epoch} to reflect optimizer overhead; all baselines from \citet{GomesZBWH25adafisher} are trained under exactly matched wall-clock budgets (200 and 90 epochs), and LLQR remains strictly within them.}
\label{tab:adafisher_cifar_imagenet}
\setlength{\tabcolsep}{3.0pt}
\renewcommand{\arraystretch}{0.92}
\scriptsize
\resizebox{\linewidth}{!}{%
\begin{tabular}{llccccccc}
\toprule
\textbf{Dataset} & \textbf{Metric}
& \textbf{SGDM}
& \shortstack{\textcolor{RoyalBlue}{\textbf{LLQR}}}
& \textbf{SGDM}\textdagger
& \textbf{Adam}\textdagger
& \textbf{AdaFisher}\textdagger
& \textbf{K-FAC}\textdagger
& \textbf{Shampoo}\textdagger\\
\midrule
\textbf{CIFAR--100} & Top-1 Acc.
& $77.52{\scriptscriptstyle \,\pm\, 0.13}$
& $\mathbf{78.34}{\scriptscriptstyle \,\pm\, 0.08}$
& $76.56{\scriptscriptstyle \,\pm\, 0.20}$
& $75.74{\scriptscriptstyle \,\pm\, 0.10}$
& $77.28{\scriptscriptstyle \,\pm\, 0.20}$
& $76.03{\scriptscriptstyle \,\pm\, 0.30}$
& $76.78{\scriptscriptstyle \,\pm\, 0.20}$\\
& Time
& $\times 1.0$
& $\times 1.25$
& $\times 1.0$
& $\times 1.16$
& $\times 1.33$
& $\times 2.32$
& $\times 10.82$\\
\midrule
\textbf{ImageNet} & Top-1 Acc.
& $77.19{\scriptscriptstyle \,\pm\, 0.02}$
& $\mathbf{77.40}{\scriptscriptstyle \,\pm\, 0.06}$
& --
& $67.78$
& $76.95$
& $70.96$
& $72.82$\\
(90 epochs) & Time
& $\times 1.0$
& $\times 1.033$
& --
& --
& --
& --
& --\\
\bottomrule
\end{tabular}
}
\vspace{-0.5em}
\end{table}

\subsection{Grokking Acceleration}
\begin{figure*}[t]
  \centering
  \includegraphics[width=\textwidth]{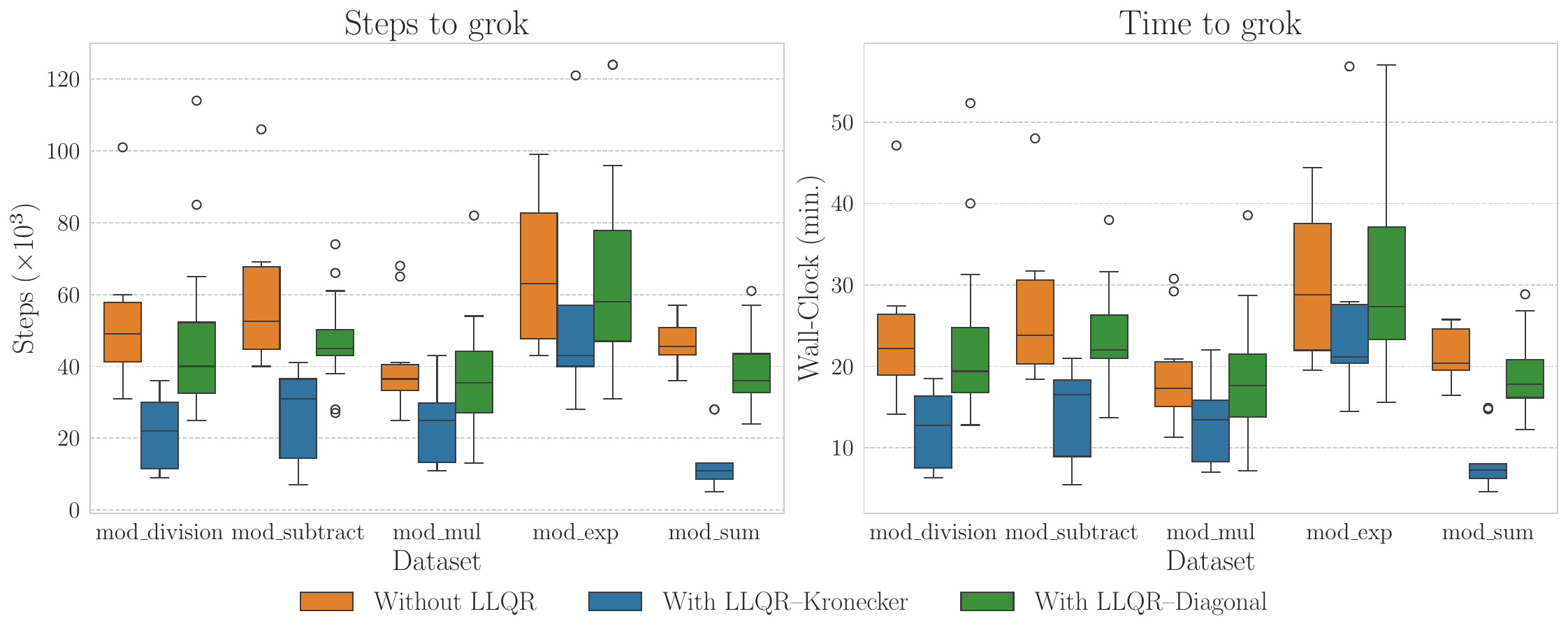}
  \caption{The grokking phenomenon---a long plateau before sudden generalization---serves as a natural testbed to assess whether LLQR accelerates learning in Transformer networks. Across five algorithmic datasets, LLQR consistently speeds up the onset of grokking in iteration count (\textbf{left}) and matches or improves wall-clock convergence (\textbf{right}).} 
  \label{fig:grokking-boxplot}
\end{figure*}

The grokking phenomenon \citep{DBLP:journals/corr/abs-2201-02177}---characterized by an extended plateau phase before sudden generalization---provides a natural testbed to evaluate whether LLQR can accelerate learning in transformer networks \citep{vaswani2017attention}. 
We tested across five common algorithmic datasets (Appendix~\ref{app:grokking_experiments_details}), with K-FAC structure and KL-induced divergence, finding that LLQR consistently accelerates the onset of grokking (reaching $\geq 95\%$ test accuracy) in terms of iteration count, and either accelerates or at least maintains convergence speed in wall-clock time (see Fig.~\ref{fig:grokking-boxplot}).
\mdfy{While the Kronecker structure again exhibits faster convergence, the diagonal structure also accelerates grokking while preserving the optimizer’s linear memory scalability---a desirable property for large transformer models.}

\subsection{IWSLT14 German to English}

\bluetext{
We further evaluate LLQR on Fairseq IWSLT14 German-to-English translation task/model \citep{DBLP:conf/wmt/OttEGA18} using the standard Transformer recipe with AdamW (Appendix~\ref{app:iwslt14_details}). As shown in Fig.~\ref{fig:imagenet_and_iwslt14_training_curves} (right) and Table~\ref{tab:imagenet_iwslt14_results}, LLQR+E-KFAC improves BLEU from $34.24\pm0.12$ to $\mathbf{34.51\pm0.05}$ with a $1.16\times$ slowdown. This setting reuses the same NGD-induced LLQR formulation and preconditioner-learning hyperparameters as the preceding large-scale experiments; the only LLQR hyperparameter changed was the preconditioner EMA decay, set to $0.925$ instead of the default $0.95$. The gain is modest but consistent across five seeds, indicating that the learned inverse preconditioner transfers to sequence-to-sequence training with little retuning.
}

%
\subsection[Autoregressive Language Modeling]{\crtext{Autoregressive Language Modeling}}
\label{sec:autoregressive_lm}

\crtext{We extend the evaluation to autoregressive language modeling with a 291M-parameter GPT-3 Small model on WikiText-103, testing whether LLQR complements both AdamW and Muon~\citep{jordan2024muon}. We use the NGD-induced quadratic with an E-KFAC structure adapted to the full word-level vocabulary (Appendix~\ref{app:wikitext103_details}). Keeping each base optimizer's hyperparameters unchanged, adding LLQR reduces best validation perplexity from $29.71$ to $29.35$ with AdamW and from $29.61$ to $29.40$ with Muon (Table~\ref{tab:wikitext103_llqr}). LLQR reaches validation perplexity below $30$ approximately one hour earlier with AdamW and $22$ minutes earlier with Muon, with respective time ratios of $1.04\times$ and $1.05\times$ through 15k updates. Applying LLQR only to Muon's auxiliary AdamW parameters yields little improvement in best perplexity. These results show that LLQR's learned geometric preconditioning can also complement Muon's momentum and orthogonalization.} 

\begin{table}[htbp]
\centering
\caption{\crtext{Effect of adding LLQR on GPT-3 Small/WikiText-103. PPL and token accuracy are validation means $\pm$ standard errors over five seeds at the checkpoint with lowest mean PPL. Threshold columns use the first evaluation with mean PPL below $30$; times are mean recorded elapsed times (h:mm). All results use the first 15k updates. Muon uses auxiliary AdamW for embeddings, normalization parameters, and biases; ``Aux.\ only'' applies LLQR only to this partition.}}
\label{tab:wikitext103_llqr}
\setlength{\tabcolsep}{4pt}
\renewcommand{\arraystretch}{1.08}
\footnotesize
\crtext{\begin{tabular}{llccccc}
\toprule
\textbf{Base} & \textbf{LLQR} & \textbf{Best PPL} & \shortstack{\textbf{Token acc.}\\\textbf{(\%)}}
& \shortstack{\textbf{Updates to}\\\textbf{PPL $<30$}} & \shortstack{\textbf{Time to}\\\textbf{PPL $<30$}} & \shortstack{\textbf{Time at}\\\textbf{15k updates}}\\
\midrule
AdamW & None & $29.71\pm0.02$ & $38.28\pm0.02$ & 11k & 6:37 & 9:07\\
AdamW & All & $\mathbf{29.35\pm0.04}$ & $\mathbf{38.50\pm0.02}$ & 9k & 5:35 & 9:27\\
\midrule
Muon & None & $29.61\pm0.04$ & $38.42\pm0.02$ & 9k & 5:23 & 9:06\\
Muon & All & $\mathbf{29.40\pm0.01}$ & $\mathbf{38.48\pm0.03}$ & 8k & 5:01 & 9:31\\
Muon & Aux.\ only & $29.59\pm0.03$ & $38.40\pm0.02$ & 9k & 5:36 & 9:28\\
\bottomrule
\end{tabular}}
\end{table}

\subsection{\mdfy{Wall-Clock Time Comparison}}
A key concern for curvature-aware and second-order methods is controlling computational cost as expressivity increases. To evaluate the practical efficiency of LLQR, we revisit CIFAR--100 with ResNet-18 under the AdaFisher setup of \citet{GomesZBWH25adafisher}, using their training configuration (Appendix~\ref{app:adafisher_training_details}). We report results over 200 SGDM+LLQR epochs, which---being faster than AdaFisher---provide a conservative comparison against the 200-epoch wall-clock budget reported for AdaFisher (Table~\ref{tab:adafisher_cifar_imagenet}). Despite using a more expressive E-KFAC block structure, rather than the diagonal approximation used by AdaFisher, LLQR yields faster training dynamics and improves final performance.

On ImageNet, we compare against the wall-clock--matched results of \citet{GomesZBWH25adafisher}, whose baselines are trained under budgets corresponding to 90 AdaFisher epochs. We train LLQR for the same 90 epochs, keeping it strictly within this budget. LLQR achieves higher top-1 accuracy than AdaFisher, despite not using the additional epochs that its lower overhead would allow. These results suggest that LLQR improves the usual fidelity--efficiency trade-off of scalable second-order methods: richer structured preconditioners can be used while maintaining overhead close to first-order training.

\section{Conclusion}

In this paper, we revisited the connection between neural network training and discrete-time optimal control. Extending prior work linking Newton steps to layerwise LQRs, we showed that a broad class of geometry-aware steepest-descent methods can be expressed as LQR problems. We then introduced a scalable relaxation that learns structured inverse preconditioners under the resulting LQR objective, avoiding explicit curvature inversion while remaining agnostic to the chosen divergence and block structure. Experiments on ResNets and Transformers show that LLQR improves convergence dynamics across settings and often yields final-performance gains with limited wall-clock overhead. \crtext{On WikiText-103, LLQR also improves best validation perplexity with both AdamW and Muon.}

{\bf Limitations:}
\bluetext{LLQR adds memory and compute overhead beyond first-order optimization, mainly from storing and periodically refitting the learned inverse preconditioner $\mU$. The cost depends on the chosen structure: Kronecker-style blocks are more expressive but costlier, while diagonal blocks are cheaper but less effective in our experiments. The preconditioner-update frequency and refitting chunk size allow practitioners to trade off wall-clock time, peak memory, and preconditioner freshness}. \crtext{The exact LQR equivalence does not ensure that finite fitting and periodic reuse of a structured preconditioner recover the exact reference update. Our autoregressive evaluation is limited to a 291M-parameter model on WikiText-103 through 15,000 updates--LLQR's performance at substantially larger pretraining scales remains untested.}

\begin{ack}
    We thank Evgenii Nikishin and Ryan D'Orazio for insightful discussions during the early stages of this project, and Damien Scieur and Mehrab Hamidi for valuable discussions during its later stages. This research was supported in part by the Canada CIFAR AI Chair program. Simon Lacoste-Julien is a CIFAR Associate Fellow in the Learning in Machines \& Brains program.
\end{ack}

\clearpage
\bibliography{lqr_paper_bib}
\bibliographystyle{plainnat}

\newpage
\appendix
\onecolumn
\section{Gradient Descent as LQR}\label{sec:gd-as-lqr}
We make explicit the correspondence between a single gradient-descent (steepest–descent under Euclidean norm) update on the parameters and a finite–horizon Linear Quadratic Regulator (LQR).
Consider a discrete dynamical system composed of $N$ layers
\[
\vx_{i+1} \;=\; f_i(\vx_i,\vtheta_i), \qquad i=0,\dots,N-1,
\]
with input $\vx_0$ (fixed during the update) and parameters $\vtheta \!=\![\vtheta_0,\dots,\vtheta_{N-1}]$.
Let the loss be $L(\vtheta):=L(\vx_N(\vtheta))$
(using the same symbol $L$ in a slight abuse of notation)
 where we assume (for clarity) that the primary dependence of $L$ on $\vtheta$ is via $\vx_N$.
At the iterate $(\vx^k,\vtheta^k)$, the gradient–descent step on $\vtheta$ solves the quadratic model:
\begin{equation}
\label{eq:sd-unconstrained}
\min_{\Delta\vtheta}\;\nabla_{\vtheta} L(\vtheta^k)^\top \Delta\vtheta \;+\; \tfrac{1}{2\eta}\sum_{i=0}^{N-1}\|\Delta\vtheta_i\|_2^2
\end{equation}

\paragraph{Linearization and sensitivity.}
Define the perturbations to current states (embeddings) and parameters as $\delta{\vx}_i \coloneqq \vx_i-\vx_i^k$ and $\delta{\vtheta}_i \coloneqq \vtheta_i-\vtheta_i^k$, with $\delta{\vx}_0=\mathbf{0}$ since the input does not vary.
Starting from the steepest–descent term
$\nabla_{\vtheta} L(\vtheta^k)^\top (\Delta\vtheta)$,
we expand by the chain rule. Because $L$ depends on $\vtheta$ through the terminal state $\vx_N$, differentiation yields:
\begin{equation}
\label{eq:chain-start}
\nabla_{\vtheta} L(\vtheta^k)^\top (\Delta\vtheta)
= \nabla_{\vx_N} L(\vx_N^k)^\top \,
\Big( \nabla_{\vtheta}\vx_N^k \cdot (\Delta\vtheta) \Big)
\end{equation}
Unrolling $\vx_N$ through the composition of layers gives
\begin{equation}
\begin{split}
\nabla_{\vtheta}\vx_N^k \cdot (\Delta\vtheta)
&= \nabla_{\vx_{N-1}} f_{N-1}^k \Big(
    \nabla_{\vx_{N-2}} f_{N-2}^k \big( \cdots
      \nabla_{\vx_{1}} f_{1}^k (\nabla_{\vtheta_0} f_0^k \,\delta\vtheta_0)
      + \nabla_{\vtheta_1} f_1^k \,\delta\vtheta_1
    \cdots \big) \\
& \hspace{6cm} + \nabla_{\vtheta_{N-1}} f_{N-1}^k \,\delta\vtheta_{N-1} \Big)
\end{split}
\end{equation}
At each layer $i$, the Jacobians $\nabla_{\vx_i} f_i^k$ and $\nabla_{\vtheta_i} f_i^k$ appear.
Collecting notation, define
\[
\mA_i \coloneqq \nabla_{\vx_i} f_i(\vx_i^k,\vtheta_i^k) = \frac{\partial f_i}{\partial \vx_i}(\vx_i^k,\vtheta_i^k),
\qquad
\mB_i \coloneqq \nabla_{\vtheta_i} f_i(\vx_i^k,\vtheta_i^k) = \frac{\partial f_i}{\partial \vtheta_i}(\vx_i^k,\vtheta_i^k)
\]

Then, propagating the perturbations forward, the deviations satisfy the linear recurrence
\begin{equation}
\label{eq:lin-dyn}
\delta{\vx}_{i+1} \;=\; \mA_i \delta{\vx}_i + \mB_i \delta{\vtheta}_i,
\end{equation}
which is the standard local linearization of each nonlinear transition.

Substituting back, the chain-rule expansion in \eqref{eq:chain-start} compactly reduces to
\begin{equation}
\label{eq:chain-terminal}
\nabla_{\vtheta} L(\vtheta^k)^\top (\Delta\vtheta)
\;=\; \nabla_{\vx_N} L(\vx_N^k)^\top \delta{\vx}_N,
\end{equation}
where $\delta{\vx}_N$ is exactly the terminal perturbation generated by the linearized dynamics \eqref{eq:lin-dyn} under controls $\delta{\vtheta}_0,\dots,\delta{\vtheta}_{N-1}$.

\paragraph{LQR form.}
Using \eqref{eq:chain-terminal} in \eqref{eq:sd-unconstrained}, the steepest–descent subproblem becomes the LQR
\begin{equation}
\label{eq:SGD_as_LQR}
\begin{aligned}
\min_{\{\delta{\vtheta}_i\}_{i=0}^{N-1}} \quad
& \nabla_{\vx_N} L(\vx_N^k)^\top\,\delta{\vx}_N
\;+\; \tfrac{1}{2}\sum_{i=0}^{N-1} \delta{\vtheta}_i^\top \delta{\vtheta}_i \\
\text{s.t.}\quad & \delta{\vx}_{i+1} = \mA_i \delta{\vx}_i + \mB_i \delta{\vtheta}_i,\;\; i=0,\dots,N-1,\\
& \delta{\vx}_0 = \mathbf{0}
\end{aligned}
\end{equation}

\paragraph{Equivalence (solution recovers the gradient).}
Let $\vlam_i$ be the costate sequence for \eqref{eq:SGD_as_LQR}. The KKT (discrete Pontryagin’s Maximum Principle) conditions 
read:
\[
\vlam_N \;=\; \nabla_{\vx_N} L(\vx_N^k),
\qquad
\vlam_i \;=\; \mA_i^\top \vlam_{i+1},\;\; i=N-1,\dots,0,
\qquad
\delta{\vtheta}_i^{\star} \;=\; -\,\eta\mI\mB_i^\top \vlam_{i+1}.
\]
Unrolling gives
\(
\vlam_{i+1} \!=\! \mA_{i+1}^\top\cdots \mA_{N-1}^\top \nabla_{\vx_N} L(\vx_N^k)
\)
and thus
\begin{equation}
\label{eq:control-equals-grad}
\delta{\vtheta}_i^{\star}
\;=\; -\,\eta\mI\mB_i^\top \mA_{i+1}^\top\!\cdots \mA_{N-1}^\top \nabla_{\vx_N} L(\vx_N^k)
\;=\; -\,\eta\nabla_{\vtheta_i} L(\vx_N^k),
\end{equation}
which is exactly the gradient component at layer \(i\). Therefore, the primal minimizer of \eqref{eq:SGD_as_LQR} yields the usual GD update
\(
\vtheta_i^{k+1} \!=\! \vtheta_i^k + \delta{\vtheta}_i^{\star}
= \vtheta_i^k - \eta \nabla_{\vtheta_i} L\big(\vx_N^k(\vtheta^k)\big).
\)

\begin{summarybox}
\textbf{Takeaway.} The linear transitions in \eqref{eq:lin-dyn}---obtained by layerwise linearization around $(\vx_i^k,\vtheta_i^k)$---are the backbone of the LQR view and do not depend on the step's metric choice. Changing the geometry (e.g., replacing the Euclidean norm in steepest descent by a different metric) alters the quadratic control penalty in \eqref{eq:SGD_as_LQR}
, \textbf{but} the linear dynamics remain the same.
\end{summarybox}

\paragraph{Remarks.}
(i) If $L$ has direct terms in $\vtheta$ (e.g., weight decay), they appear as additional stage terms in \eqref{eq:SGD_as_LQR}.
(ii) The derivation remains valid if the loss depends on intermediate states; such terms enter the cost through $\sum_i \nabla_{\vx_i} L^\top \delta{\vx}_i$.
(iii) Solving \eqref{eq:SGD_as_LQR} via the Riccati recursion reproduces backpropagation algebraically; cf.\ the costate recursion above.

\section{Generic Steepest Descent as LQR}\label{app:steepest-descent-as-lqr-detailed}

We now extend the Euclidean case of Appendix~\ref{sec:gd-as-lqr} to general steepest descent.
The derivation follows \citet[Ch.~2.6]{bertsekas2016nonlinear}, but generalizes beyond Newton’s method to a broader family of divergence-induced quadratic models.
The key observation is that only the quadratic penalty on parameter perturbations changes: the local linear dynamics and sensitivity relations remain the same as in \eqref{eq:lin-dyn}--\eqref{eq:chain-terminal}.

\subsection{Setup and notation}
Consider the $N$-layer system $\vx_{i+1}=f_i(\vx_i,\vtheta_i)$ with fixed input $\vx_0$ and parameters $\vtheta=[\vtheta_0,\dots,\vtheta_{N-1}]$.
Let the forward map
\[
\vx(\vtheta)\;\coloneqq\;[\vx_1(\vtheta),\vx_2(\vtheta),\ldots,\vx_N(\vtheta)]^\top,
\]
so that the loss writes $L(\vtheta):=L(\vx_N(\vtheta))$ (primary dependence via $\vx_N$, the network's output, for clarity. By a slight abuse of notation, we use the same symbol $L$ for the functions of the parameter $\theta$ on the left-hand side and the function of the activations $\vx$ on the right-hand side).
Again, iterate $(\vx^k,\vtheta^k)$ define perturbations
\[
\delta{\vx}_i\coloneqq \vx_i-\vx_i^k,\qquad
\delta{\vtheta}_i\coloneqq \vtheta_i-\vtheta_i^k,\qquad
\delta{\vx}_0=\mathbf{0}
\]

\subsection{Problem restatement}
Under a  divergence~\citep{amari2000methods} 
$D$ with $D(\vz,\vz)=0$ and $D(\vz,\vw)\ge 0$, we want to solve the steepest-descent step at $(\vx^k,\vtheta^k)$:
\begin{equation}
\label{eq:app-sd-div}
\min_{\delta{\vtheta}}\;
\nabla_{\vtheta} L\big(\vx(\vtheta^k)\big)^\top\,\delta{\vtheta}
\;+\;
\tfrac{1}{2}\,\delta{\vtheta}^\top
\big(\nabla^2_{\vtheta\vtheta}D(\vtheta^k,\vtheta)\big|_{\vtheta=\vtheta^k}\big)\delta{\vtheta},
\end{equation}
where $\vx(\vtheta)$ is the forward map producing the terminal state $\vx_N$ and $D$ is layerwise decomposable, i.e.\
$D(\vtheta^k,\vtheta) = \psi_N(\vx_N(\vtheta)) + \sum_{i=0}^{N-1} \psi_i(\vx_i(\vtheta),\vtheta_i)$ for some twice differentiable functions $\psi_i, \, i=0\cdots N$.

As in Appendix~\ref{sec:gd-as-lqr}, the perturbations obey the linearized dynamics
\[
\delta{\vx}_{i+1} \;=\; \mA_i \delta{\vx}_i + \mB_i \delta{\vtheta}_i,
\qquad
\mA_i=\nabla_{\vx_i}f_i(\vx_i^k,\vtheta_i^k),
\qquad
\mB_i=\nabla_{\vtheta_i}f_i(\vx_i^k,\vtheta_i^k),
\]
with $\delta{\vx}_0=\mathbf{0}$.

\textbf{Note:} The requirement that $D$ be layerwise decomposable is not restrictive. If a term $\psi_i$ depends directly on another state $\vx_j$ (or parameter $\vtheta_j$), this can be handled by augmenting $\vx_i$ with $\vx_j$ (or $\vtheta_j$). In practice, this is equivalent to introducing skip connections in the network.
However, note that this class of divergence-induced models covers nearly all second-order methods of interest. For example,
\begin{itemize}
\item {\bf (Damped) Newton} \label{app:damped_Newton}  is obtained by considering the Bregman gap of the locally regularized loss:
\[
D_{\text{N}}(\vtheta^{(k)},\vtheta)
:= F(\vtheta) - F(\vtheta^{(k)}) - \nabla F(\vtheta^{(k)})^\top (\vtheta-\vtheta^{(k)}), \quad F(\vtheta) := L(\vtheta) + \frac{\lambda}{2}(\vtheta - \vtheta^k)^2
\]
with $\lambda >0$ chosen so that $F$ is locally strictly convex in a convex neighborhood of $\theta^k$. $D_N$ takes the layerwise form \eqref{eq:layerwise} and  $\mH = \nabla^2L + \lambda _id$.

\item {\bf Gauss-Newton.} Let the sample loss be $\ell(\vx_N) = \tfrac12 \|\vr(\vx_N)\|^2$ with residual $\vr$.
Define the output-space divergence
\[
D_{\text{GN}}(\vx_N^k,\vx_N) :=
\tfrac12 (\vx_N-\vx_N^k)^\top \mW(\vx_N^k)(\vx_N-\vx_N^k),
\quad
\mW(\vx_N^k) := J_{\vr}(\vx_N^k)^\top J_{\vr}(\vx_N^k).
\]
Pulling this divergence back to parameter space via the network Jacobian
$J_{\vtheta}\vx_N$ yields the quadratic model
\[
m(\Delta\vtheta) = \vg^\top \Delta\vtheta
+ \tfrac12 \Delta\vtheta^\top (J_{\vtheta}\vx_N^\top \mW J_{\vtheta}\vx_N)\,\Delta\vtheta,
\]
so that $\mH=J_{\vtheta}\vx_N^\top \mW J_{\vtheta}\vx_N$, i.e.\ the Gauss--Newton matrix.
\item

\textbf{Intermediate-layer metrics.}
Include layer terms
\[
D_{\text{int}} := \sum_{i\in\mathcal K}
\tfrac12(\vx_i(\vtheta)-\vx_i(\vtheta^{(k)}))^\top \mW_i(\vx_i(\vtheta^{(k)}))
(\vx_i(\vtheta)-\vx_i(\vtheta^{(k)}))
\]
so $\mH$ pulls back these activation metrics via the chain rule.

\item \textbf{Natural gradient / Fisher.} \label{app:NGD_Fisher_from_KL}
For probabilistic models, let $p_{\vtheta} (y\mid \vx_N)$ and define
\[
D_{\text{NG}}(\vtheta^{(k)},\vtheta)
:= \mathrm{KL}\big(p_{\vtheta^{(k)}}(\cdot\mid \vx_N(\vtheta^{(k)}))
\;\|\; p_{\vtheta}(\cdot\mid \vx_N(\vtheta))\big)
\]
linearized at $\vtheta^{(k)}$.
Its Hessian at $\vtheta^{(k)}$ is the Fisher information matrix.

\end{itemize}

\subsection{Layerwise decomposition of the local quadratic model}

Introducing the block-constraint vector (one row per layer transition)
\[
h(\vx,\vtheta)\;\coloneqq\;
\begin{bmatrix}
f_0(\vx_0,\vtheta_0)-\vx_1\\
\vdots\\
f_{N-1}(\vx_{N-1},\vtheta_{N-1})-\vx_N
\end{bmatrix}
=\mathbf{0}
\quad\text{on feasible trajectories.}
\]
Define $\mathcal{H}$ (PMP-style) \citep{pontriagin1962mathematical}
\begin{equation}
\label{eq:app-H}
\mathcal{H}(\vx,\vtheta,\vp)
= \psi_N(\vx_N)
+ \sum_{i=0}^{N-1} \Big[\psi_i(\vx_i,\vtheta_i)
+ \vp_{i+1}^\top\!\big(f_i(\vx_i,\vtheta_i)-\vx_{i+1}\big)\Big],
\end{equation}
where $\vp=[\vp_1,\ldots,\vp_N]$ stacks Lagrange multipliers (costates) associated with the state constraints.

Because $h=0$ along feasible trajectories, we have the identity
\begin{equation}
\label{eq:app-D-equals-H}
D(\vtheta^k,\vtheta)=\mathcal{H}(\vx(\vtheta),\vtheta,\vp)
\quad\text{for any }\vp.
\end{equation}
We may select $\vp$ so that the following stationarity condition w.r.t.\ states holds:
\begin{equation}
\label{eq:app-Hx-0}
\nabla_{\vx} \mathcal{H}\big(\vx(\vtheta^k),\vtheta^k,\vp\big) \;=\; 0,
\end{equation}
which yields the discrete adjoint recursion (componentwise, using the layerwise decomposition of $D$):
\begin{equation}
\label{eq:app-adjoint}
\vp_N \;=\; \nabla_{\vx_N} \psi_N(\vx_N^k),
\qquad
\vp_i \;=\; \mA_i^\top \vp_{i+1} + \nabla_{\vx_i} \psi_i(\vx_i^k,\vtheta_i^k),
\quad i=N-1,\ldots,1.
\end{equation}

\subsection[From quadratic divergence to an LQR cost]{From $\delta{\vtheta}^\top
   \big(\nabla^2_{\vtheta\vtheta}D(\vtheta^k,\vtheta)\big|_{\vtheta=\vtheta^k}\big)\,\delta{\vtheta}$ to a quadratic LQR cost}
Differentiate \(\mathcal{H}(\vx(\vtheta),\vtheta,\vp)\) w.r.t.\ \(\vtheta\) using the chain rule:
\begin{equation}
\label{eq:app-dthetaH}
\nabla_{\vtheta}\big[\mathcal{H}(\vx(\vtheta),\vtheta,\vp)\big]
= \big(\nabla_{\vtheta}\vx(\vtheta)\big)^\top \nabla_{\vx} \mathcal{H}(\vx(\vtheta),\vtheta,\vp)
\;+\; \nabla_{\vtheta} \mathcal{H}(\vx(\vtheta),\vtheta,\vp).
\end{equation}
By the choice \eqref{eq:app-Hx-0}, the first term vanishes at \(\vtheta^k\), hence
\[
\nabla_{\vtheta} D(\vtheta^k,\vtheta)\big|_{\vtheta=\vtheta^k}
= \nabla_{\vtheta} \mathcal{H}(\vx(\vtheta^k),\vtheta^k,\vp).
\]
Now differentiate \eqref{eq:app-dthetaH} again at \(\vtheta^k\) in the direction \(\delta{\vtheta}\) to obtain the quadratic form:
\begin{align}
\delta{\vtheta}^\top \nabla_{\vtheta\vtheta}^2 D(\vtheta^k,\vtheta^k)\big|_{\vtheta=\vtheta^k}\, \delta{\vtheta}
&= \delta{\vtheta}^\top
\Big[
(\nabla_{\vtheta}\vx)^\top \nabla_{\vx\vx}^2 \mathcal{H} \, (\nabla_{\vtheta}\vx)
+ (\nabla_{\vtheta}\vx)^\top \nabla_{\vx\vtheta}^2 \mathcal{H}
+ \nabla_{\vtheta\vx}^2 \mathcal{H} \, (\nabla_{\vtheta}\vx)
+ \nabla_{\vtheta\vtheta}^2 \mathcal{H}
\Big] \delta{\vtheta} \nonumber\\
&\quad +\;\underbrace{\delta{\vtheta}^\top \big[\nabla_{\vtheta\vtheta}^2 \vx\big]^\top \nabla_{\vx} \mathcal{H} \,\delta{\vtheta}}_{=\,0\ \text{by }\eqref{eq:app-Hx-0}}.
\label{eq:app-hess-expand}
\end{align}
Remarking that deriving the forward map gives you the backward map, \ \(\delta{\vx} = (\nabla_{\vtheta} \vx)\,\delta{\vtheta}\), since by recursion:
\begin{align*}
    \nabla_{\vtheta} \vx_i\,\delta{\vtheta}_i &= (\nabla_{\vx}f_i(\nabla_{\theta} \vx_{i-1}) + \nabla_{\vtheta}f_{i}) \,\delta{\vtheta}_i \\
    &= \mA_i(\mA_{i-1}(\hdots) + \mB_{i-1})\delta{\vtheta}_i + \mB_i\delta{\vtheta}_i \\
    &= \delta{\vx}_i
\end{align*}
Each block becomes
\begin{equation}
\label{eq:app-blocks}
\delta{\vtheta}^\top (\nabla_{\vtheta}\vx)^\top \nabla_{\vx\vx}^2 \mathcal{H} (\nabla_{\vtheta}\vx)\, \delta{\vtheta}
= \delta{\vx}^\top \nabla_{\vx\vx}^2 \mathcal{H} \,\delta{\vx},\qquad
\delta{\vtheta}^\top (\nabla_{\vtheta}\vx)^\top \nabla_{\vx\vtheta}^2 \mathcal{H} \,\delta{\vtheta}
= \delta{\vx}^\top \nabla_{\vx\vtheta}^2 \mathcal{H} \,\delta{\vtheta},
\end{equation}
and similarly for the mixed term with \(\nabla_{\vtheta\vx}^2 \mathcal{H}\). Moreover, from having $D \in \mathcal{C}^2$ and $f \in \mathcal{C}^2$, $\nabla_{\vx\vtheta}^2 \mathcal{H}$ is a symmetric matrix ($\nabla_{\vx\vtheta}^2 \mathcal{H} = \nabla_{\vtheta\vx}^2 \mathcal{H}$)
Hence \eqref{eq:app-hess-expand} gives the desired decomposition:
\begin{equation}
\label{eq:app-decomp}
\delta{\vtheta}^\top \nabla_{\vtheta\vtheta}^2 D(\vtheta^k,\vtheta)\big|_{\vtheta=\vtheta^k}\, \delta{\vtheta}
=
\delta{\vx}^\top \nabla_{\vx\vx}^2 \mathcal{H}\, \delta{\vx}
+ 2\delta{\vtheta}^\top \nabla_{\vtheta\vx}^2 \mathcal{H}\, \delta{\vx}
+ \delta{\vtheta}^\top \nabla_{\vtheta\vtheta}^2 \mathcal{H}\, \delta{\vtheta}.
\end{equation}

\paragraph{LQR form.}
Combining \eqref{eq:app-decomp} with the linear term
\(\nabla_{\vx_N} L(\vx^k,\vtheta^k)^\top\,\delta{\vx}_N\)
(from \eqref{eq:chain-terminal}), and the linearized dynamics \eqref{eq:lin-dyn}, we obtain the LQR: 
\begin{align*}
\min_{\{\delta{\vtheta}_i\}_{i=0}^{N-1}} \quad
& \nabla_{\vx_N} L(\vx^k,\vtheta^k)^\top\,\delta{\vx}_N +\tfrac{1}{2}\delta\vx_N^\top\nabla_{\vx_N\vx_N}^2 \psi_N(\vx_N)\delta\vx_N  \\
& \qquad\;+\tfrac{1}{2}\sum_{i=0}^{N-1}\!\big[
\delta{\vx}_i^\top \nabla_{\vx_i\vx_i}^2 \mathcal{H}_i\, \delta{\vx}_i
+ 2\delta{\vtheta}_i^\top \nabla_{\vtheta_i\vx_i}^2 \mathcal{H}_i\, \delta{\vx}_i
+ \delta{\vtheta}_i^\top \nabla_{\vtheta_i\vtheta_i}^2 \mathcal{H}_i\, \delta{\vtheta}_i
\big]\\ 
\text{s.t.}\quad
 &\delta{\vx}_{i+1}=\mA_i\delta{\vx}_i+\mB_i\delta{\vtheta}_i, \, \delta{\vx}_0=\mathbf{0}.
\end{align*}

\subsection{Riccati solution (closed form)}
\label{secapp:riccati}
Let the block matrices be
\[
\mQ_i=\nabla^2_{\vx_i\vx_i}\mathcal{H}_i,\quad
\mR_i=\nabla^2_{\vtheta_i\vtheta_i}\mathcal{H}_i,\quad
\mM_i=\nabla^2_{\vtheta_i\vx_i}\mathcal{H}_i,
\]
and terminal data \(\mQ_N=\nabla_{\vx_N\vx_N}^2 \psi_N(\vx_N)\), \(\va_N=\nabla_{\vx_N}L(\vx^k,\vtheta^k)\).
The finite-horizon LQR is solved by the backward Riccati recursion \citep{kalman1960contributions,anderson2007optimal}:
\begin{align*}
\mK_i &= \mA_i^\top \mK_{i+1}\mA_i + \mQ_i
- (\mA_i^\top\mK_{i+1}\mB_i + \mM_i^\top)
(\mR_i + \mB_i^\top\mK_{i+1}\mB_i)^{-1}
(\mM_i + \mB_i^\top\mK_{i+1}\mA_i), \\
\vlambda_i &= \mA_i^\top \vlambda_{i+1}
- (\mA_i^\top\mK_{i+1}\mB_i + \mM_i^\top)
(\mR_i + \mB_i^\top\mK_{i+1}\mB_i)^{-1}
\mB_i^\top \vlambda_{i+1},
\end{align*}
with terminal conditions \(\mK_N=\mQ_N\), \(\vlambda_N=\va_N\).
The optimal updates follow:
\begin{align*}
&\delta{\vtheta}_i^{\star}
   = -(\mR_i + \mB_i^\top \mK_{i+1}\mB_i)^{-1}
      \Big[(\mM_i+\mB_i^\top\mK_{i+1}\mA_i)\delta{\vx}_i^{\star}
          + \mB_i^\top \vlambda_{i+1}\Big] \\
&\delta{\vx}_{i+1}^{\star} = \mA_i \delta{\vx}_i^{\star} + \mB_i \delta{\vtheta}_i^{\star}
\end{align*}

\begin{summarybox}
\paragraph{Takeaway.}
Exactly as in Appendix~\ref{sec:gd-as-lqr}, the \textit{dynamics} \(\delta{\vx}_{i+1}=\mA_i\delta{\vx}_i+\mB_i\delta{\vtheta}_i\) are fixed by the network; the \textit{geometry} (from divergence \(D\)) appears only through the quadratic cost blocks \((\mQ_i,\mR_i,\mM_i)\). Solving the LQR via the Riccati equations yields a preconditioned steepest–descent update under that geometry.
\end{summarybox}

\section{Natural Gradient Descent}\label{app:natural-gradient-as-steepest-descent}
Natural Gradient Descent (NGD) \citep{Amari1998NaturalGW} follows the steepest-descent path in a space scaled by Fisher's information matrix ($\mF$):
\begin{equation}\label{eq:Steepest_descent_under_Fisher}
        \vd^{\star} = \argmin_{\vd \in \mathbb{R}^n}
\left\{ \nabla f(\vx^k, \vtheta^k)^\top \vd + \tfrac{1}{2}\vd^\top \mF \vd \right\}.
\end{equation}
Leveraging the well-known relationship \citep{amari2000methods, DBLP:journals/jmlr/Martens20}\footnote{See~\url{https://jaketae.github.io/study/natural-gradient} for a straightforward derivation.} between Fisher's matrix and Kullback–Leibler divergence:
\begin{equation*}
    \mF = \nabla^2_{\vtheta'\vtheta'} \KL[p(\vx|\vtheta)||p(\vx|\vtheta')]
\end{equation*}
Allows us to rewrite \eqref{eq:Steepest_descent_under_Fisher}:
\begin{equation}\label{eq:Steepest_descent_under_KL}
\begin{aligned}
        \vd^{\star} = \argmin_{\vd \in \mathbb{R}^n}\:&\nabla f(\vx^k, \vtheta^k)^\top \vd \\
&\,+ \tfrac{1}{2}\vd^\top \nabla^2_{\vtheta'\vtheta'} \KL[p(\vx|\vtheta)||p(\vx|\vtheta')] \vd \\
\end{aligned}
\end{equation}
Such that we can directly apply theorem~\ref{theo:LLQR} to recover an \textbf{exact} NGD update by solving the associated LQR.
\begin{lemma}\label{lem:ngd-equivalent-to-lqr}
    The NGD update $\delta{\vtheta}$ at iteration $k$ can be recovered by solving:
    \begin{equation}\label{eq:LQR_equivalent_to_NGD}
    \begin{split}
        \min_{\delta{\vtheta}} \enspace  &\vec{a}_N^T \delta\vx_N + \frac{1}{2}\delta\vx_N^T\mQ_N\delta\vx_N + \sum_{i=0}^{N-1} \big( \frac{1}{2}\delta\vx_i^T \mQ_i \delta\vx_i + \frac{1}{2}\delta{\vtheta}_i^T \mR_i \delta{\vtheta}_i + \delta{\vtheta}_i^T \mM_i \delta\vx_i \big)\\
        \mathrm{s.t} \enspace & \delta{\vx}_{i+1} = \mA_i\delta{\vx}_i + \mB_i\delta{\vtheta}_i\\
        & \delta{\vx}_0 = 0
    \end{split}
\end{equation}
where
\begin{gather*}
    \mQ_N=\nabla^2_{\vx'_N} \KL(p_{\vx_N}||p_{\vx'_N})\Big|_{\vx'_N =\vx_N}\: ;\: \mQ_i = \nabla^2_{\vx_i \vx_i}\mathcal{H}_i, \\
    \vec{a}_N = \nabla_{\vx_N} L(\vx^k,\vtheta^k)^{\!\top}\: ;\: \mR_i=\nabla^2_{\vtheta_i \vtheta_i}\mathcal{H}_i\: ;\: \mM_i=\nabla^2_{\vtheta_i \x_i}\mathcal{H}_i
\end{gather*}
for which an analytical solution, given by Riccati's equation, exists:
\begin{equation}\label{eq:Riccati_update_steepest_descent}
\begin{aligned}
    & \delta{\vtheta}^\star_0 = -(\mR_0 + \mB_0^T \mK_1\mB_0)^{-1} \mB_0^T \vlambda_1 \\
    & \delta{\x}^\star_{i+1} = \mA_i\delta{\vx}^\star_i + \mB_i\delta{\vtheta}^\star_i \\
    & \delta{\vtheta}^\star_i = -(\mR_i + \mB_i^T \mK_{i+1}\mB_i)^{-1}\left(( \mM_i +\mB_i^T\mK_{i+1}\mA_i)\delta{\x}^\star_i + \mB_i^T\vlambda_{i+1}\right) \\
    & \mK_N = \mQ_N \; ; \; \vlambda_N= \va_N \\
    &\mK_i = \mA^T_i\mK_{i+1}\mA_i + \mQ_i - (\mA^T_i\mK_{i+1}\mB_i + \mM_i^T)(\mR_i + \mB_i^T\mK_{i+1}\mB_i)^{-1}(\mM_i + \mB_i^T\mK_{i+1}\mA_i) \\
    &\vlambda_i = \mA^T_i\vlambda_{i+1} - (\mA^T_i\mK_{i+1}\mB_i + \mM_i^T)(\mR_i + \mB_i^T\mK_{i+1}\mB_i)^{-1}\mB_i^T\vlambda_{i+1}
\end{aligned}
\end{equation}

\end{lemma}

\section{\mdfy{Further Experiments}}\label{app:further_exps}
\subsection{Rosenbrock Toy Experiment}
\begin{figure*}[tb]
  \centering
  \includegraphics[width=\textwidth]{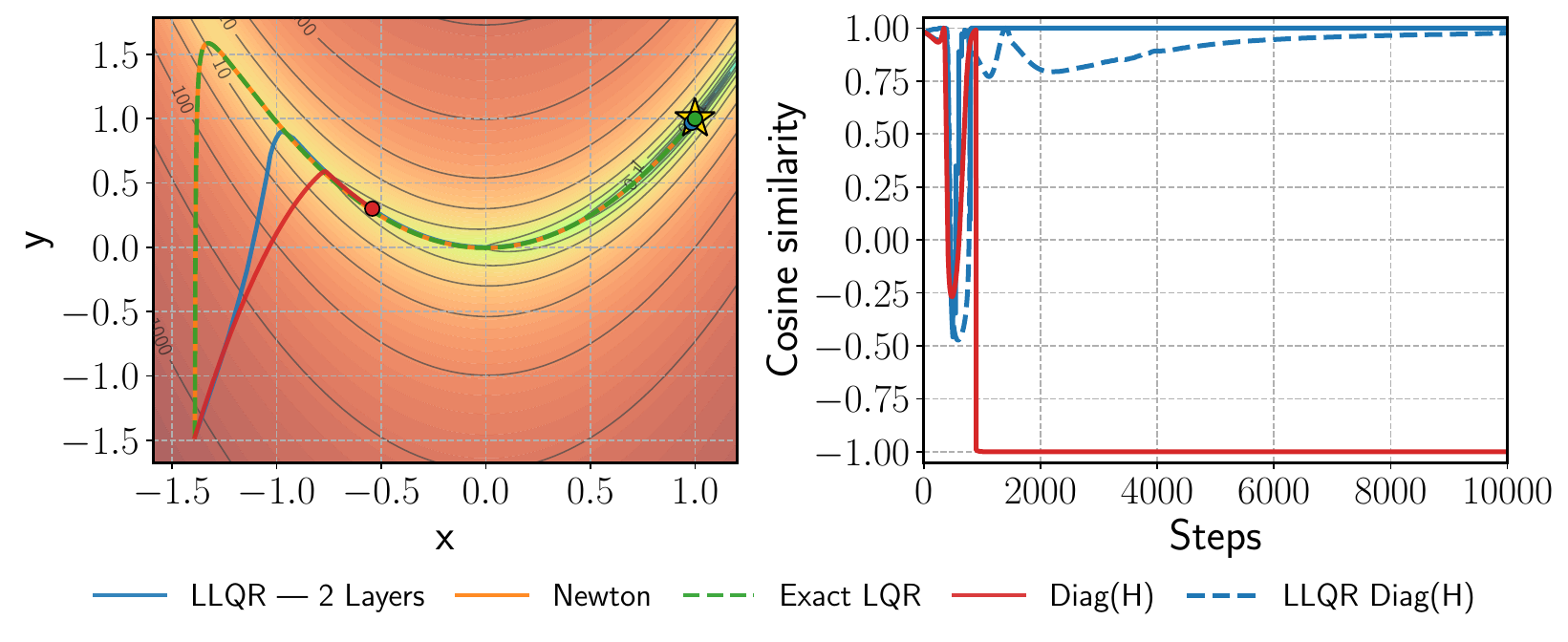}
  \caption{\textbf{Rosenbrock extended diagnostics.}
  \textbf{Left:} optimization trajectories over Rosenbrock level curves for Newton's method, exact LQR, two-layer LLQR, and a diagonal-Hessian Newton approximation.
  \textbf{Right:} cosine similarity between each approximate update direction and the exact Newton direction $-\mH^{-1}g$.
  Despite using diagonal degrees of freedom, LLQR learns an update-aligned correction that remains close to Newton, whereas $\operatorname{diag}(\mH)^{-1}$ fails to produce a reliably aligned update by discarding cross-layer curvature couplings. The dotted blue line indicates cosine similarity between $-\operatorname{diag}(\mH)^{-1}g$ and $-\mH^{-1}g$ taken along the LLQR path.}
  \label{fig:rosenbrock_diagnostics}
\end{figure*}

\bluetext{We supplement the Rosenbrock experiment with trajectory visualizations and directional diagnostics against the exact Newton step. Fig.~\ref{fig:rosenbrock_diagnostics} compares Newton's method, exact LQR, a two-layer LLQR preconditioner, and the diagonal-Hessian Newton approximation. Although LLQR uses only diagonal degrees of freedom in this experiment, learning them dynamically through the LLQR relaxed objective yields an \textbf{update-aligned curvature correction}: LLQR remains close to the Newton and exact-LQR trajectories, while preserving the effect of inter-layer couplings that $\operatorname{diag}(\mH)$ discards before inversion. This difference is also reflected in the cosine similarity with the Newton direction, where LLQR remains substantially better aligned than the diagonal-Hessian baseline throughout the Rosenbrock valley.}

\subsection[Coupled objective and block-diagonal ablation]{\crtext{Coupled objective and block-diagonal ablation}}
\label{app:coupled-objective-ablation}

\begin{figure}[htbp]
\centering
\includegraphics[width=\textwidth]{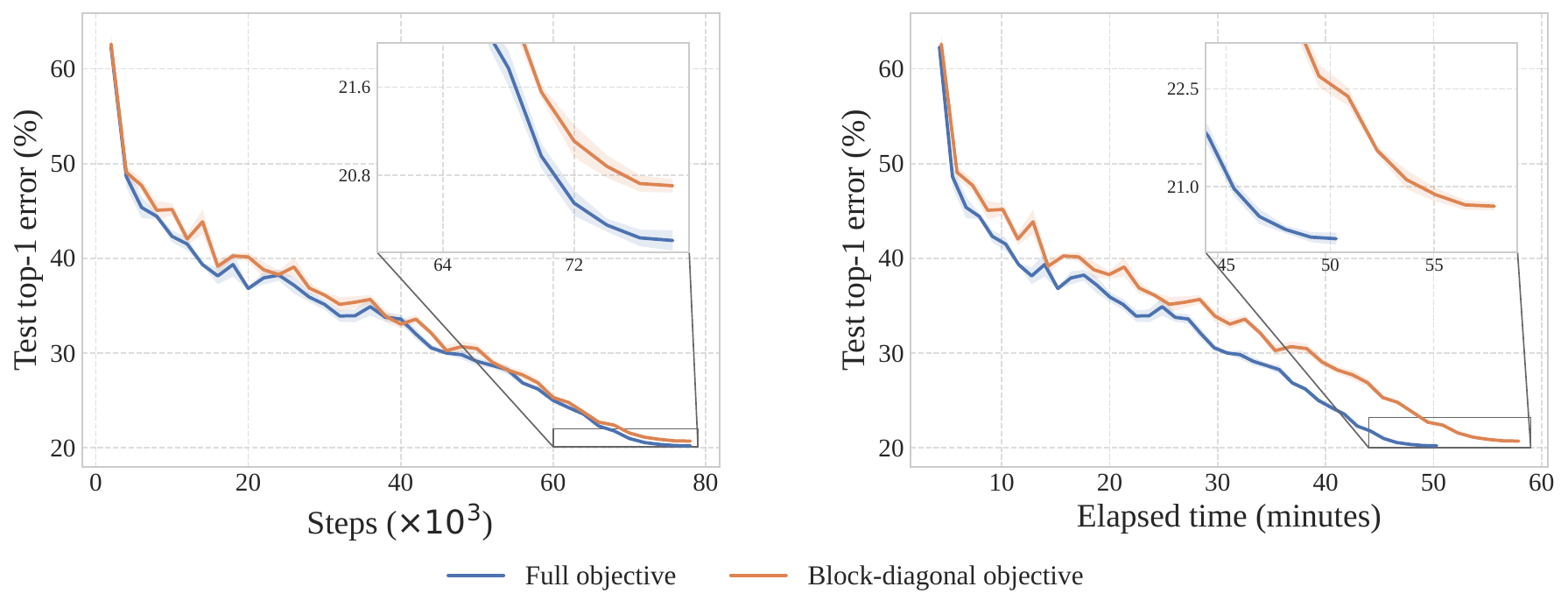}
\caption{\crtext{Full versus block-diagonal fitting objectives for Newton/E-KFAC LLQR on ResNet-18/CIFAR--100. Test top-1 error is plotted against updates (left) and recorded elapsed time (right). Results are averaged over five seeds, with shaded regions indicating standard errors; all experiments were run on NVIDIA L40S GPUs.}}
\label{fig:block-diagonal-ablation}
\end{figure}

\crtext{At a fixed iterate and minibatch, let $\vg=\nabla L(\vtheta^k)$ and let $\mH$ be the curvature in Eq.~\eqref{eq:div_hessian}. Substituting $\Delta\vtheta=-\mU\vg$ into the quadratic model gives}
\begin{equation}
J_{\mathrm{full}}(\mU)
=-\vg^\top\mU\vg
+\tfrac12(\mU\vg)^\top\mH(\mU\vg).
\label{eq:full-fitting-objective}
\end{equation}
\crtext{This is the same objective evaluated by the layerwise formulation. The structure imposed on $\mU$ restricts the update family, but leaves cross-block curvature in the fitting objective.}

\crtext{To isolate these couplings, we retain the parameter partition of $\mU=\operatorname{diag}(\mU_i)$ and instead fit}
\begin{equation}
J_{\mathrm{block}}(\mU)
=-\sum_i\vg_i^\top\mU_i\vg_i
+\tfrac12\sum_i(\mU_i\vg_i)^\top\mH_{ii}(\mU_i\vg_i).
\label{eq:block-fitting-objective}
\end{equation}
\crtext{Each $\mH_{ii}$ retains the complete downstream curvature for parameter block $i$; only cross-block terms are removed. For Newton's quadratic $C(\vu)=\vg^\top\vu+\tfrac12\vu^\top\mH\vu$, let $\vu=-\mU\vg$ and let $\Pi_i$ retain only control block $i$. Then}
\begin{equation}
\sum_i C(\Pi_i\vu)
=\sum_i\vg_i^\top\vu_i+\tfrac12\sum_i\vu_i^\top\mH_{ii}\vu_i
=J_{\mathrm{block}}(\mU).
\label{eq:projected-control-ablation}
\end{equation}
\crtext{We evaluate each projected control through the unchanged LLQR dynamics, requiring one rollout per controlled parameter block without forming Hessian blocks.}

\crtext{Figure~\ref{fig:block-diagonal-ablation} compares the two objectives on ResNet-18/CIFAR--100 with Newton-induced curvature and the same E-KFAC parameterization. Both use five matched initialization seeds, refits every 250 updates, and 25 inner SGDM steps; remaining settings follow Appendix~\ref{app:CIFAR_experiments_details}. Retaining cross-block terms improves mean per-seed best test accuracy by $0.52\pm0.09$ percentage points, with gains on all five seeds and the clearest separation late in training.}

\subsection{ResNets}
\begin{figure*}[tb]
  \centering
  \includegraphics[width=\textwidth]{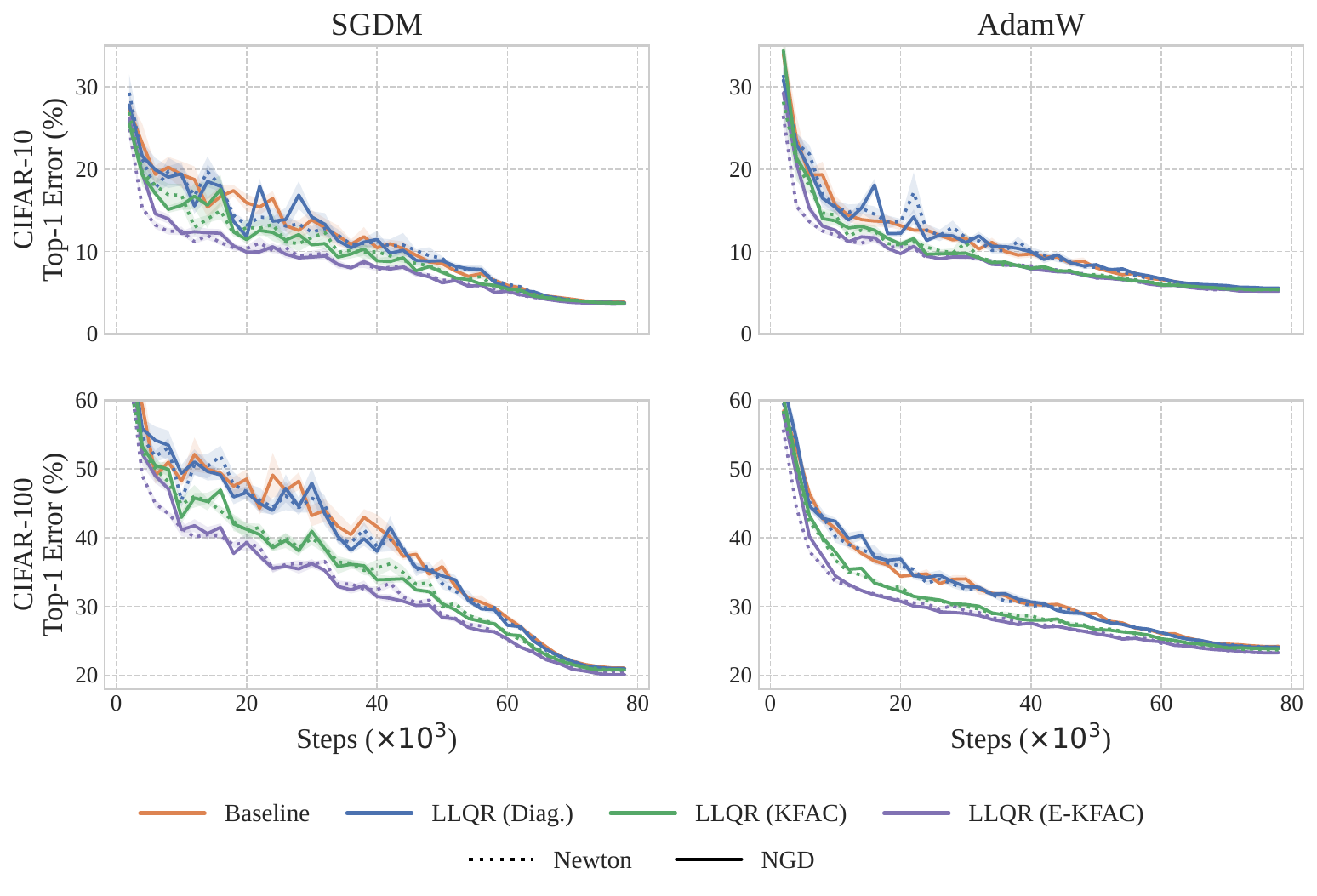}
  \caption{ResNet-18 training curves on CIFAR--10 and CIFAR--100 with SGDM and AdamW under state-of-the-art training settings. Rows separate datasets and columns separate optimizers; curves report Top-1 error versus training steps for the baseline and LLQR variants using diagonal, K-FAC, and E-KFAC block structures under Newton- and NGD-induced quadratics. Results are averaged over five seeds, with shaded regions indicating standard errors; all experiments were run on NVIDIA L40S GPUs.}
  \label{fig:cifar10_cifar100_top1_error_curves}
\end{figure*}

In addition to Fig.~\ref{fig:imagenet_and_iwslt14_training_curves} in the main paper, we present below similar curves for ResNet-18 on CIFAR--10/100, for both SGDM and AdamW. The training behaviour remains the same across those experiments (Fig.~\ref{fig:cifar10_cifar100_top1_error_curves}).

\section{Algorithmic Details}\label{app:algo-details}
\subsection{Damping}
Methods such as K-FAC incorporate damping to regularize the preconditioner $(\bar{\mP} = \mP + \mI)$---similar to Tikhonov regularization \citep{Tikhonov1943OnTS}---to ensure stability when dealing with near-zero eigenvalues. In our method, damping is conveniently incorporated into the $\mR_i$ matrices ($\bar{\mR}_i = \mR_i + \lambda_i\mI$), equivalent to imposing a trust-region of the form $\lambda_i/2 ||\delta{\vtheta}_i||$ 
over the update in the LQR objective (Eq.~\eqref{eq:relaxed-lqr-equivalence}). This per-layer damping can also help to stabilize the inversion step $(\bar{\mR_i} + \mB_i^T\mK_{i+1}\mB_i)^{-1} \to (\lambda_i\mI + \mR_i + \mB_i^T\mK_{i+1}\mB_i)^{-1}$ involved in the Riccati solutions (Eq.~\eqref{eq:Riccati_update_steepest_descent}).


\subsection{Efficient implementation}
All matrix operations in our formulation are implemented using automatic differentiation primitives—Jacobian--vector products (JVPs), vector--Jacobian products (VJPs), and Hessian--vector products (HVPs)—without explicitly forming or inverting large matrices. In particular, the matrices $Q_i$, $R_i$, $M_i$, as well as all Jacobians involved in the updates, are never instantiated. Instead, all required quantities are computed on the fly via JVPs, VJPs, and HVPs, pre-compiled over each layer’s parameters and representations.

The memory cost of these operations scales \textbf{linearly} with the dimension of the layer’s representation or parameter space, and the associated intermediate buffers are released immediately after each preconditioner update. As a result, the overall memory footprint remains controlled and comparable to that of standard backpropagation. These techniques are well supported by modern automatic differentiation frameworks \citep{DBLP:journals/neco/Pearlmutter94, DBLP:journals/jmlr/BaydinPRS17} and form the basis of efficient implementations of second-order and curvature-aware optimization methods \citep{DBLP:conf/icml/Martens10, DBLP:conf/aistats/MengVL0L20}.

\subsubsection{\bluetext{Chunking}}\label{app:chunking}
When the learned preconditioner is refit on a large minibatch, the batch
dimension can be treated as an implementation axis rather than as an additional
modeling approximation. In our implementation, we allow to partition the
preconditioner-update batch into smaller chunks and evaluates the same relaxed
LLQR objective on each chunk, with the loss-gradient and preconditioner-gradient
contributions accumulated using the corresponding chunk weights. Thus the
effective objective remains the minibatch objective used to update $\mU$, while
the largest activation, adjoint, and second-order intermediate tensors scale
with the chunk size rather than with the full preconditioner batch. This yields
a practical memory--compute trade-off: smaller chunks reduce peak memory and
make larger preconditioner batches feasible, at the cost of additional repeated
JVP/VJP/HVP work and less full-batch fusion.

The chunked update exploits the particular block structure of the batched LQR
operators. For a minibatch, the state variables $\delta\vx_i$ are
batch-indexed, so the state-state curvature block $\mQ_i$ is block-structured
over examples, whereas the control $\delta\vtheta_i$ is shared across the
batch and the mixed block $\mM_i$ acts as a block column mapping
batch-indexed state perturbations to a single parameter-side contribution. The
implementation therefore applies the $\mathcal{H}$ actions needed by the relaxed
objective chunk by chunk: parameter-side terms such as
$\mR_i\delta\vtheta_i+\mM_i\delta\vx_i$ are reduced immediately into the
shared parameter space, while state-side terms such as
$\mM_i^\top\delta\vtheta_i+\mQ_i\delta\vx_i$ are produced only for the current
chunk. The same idea is used at the grouped LLQR-segment level, so several
execution stages can be traversed as one segment while retaining only the
segment boundary state needed by the LQR recursion. This keeps the
implementation compatible with the exact mixed second-order actions used by
LLQR, while exposing the chunk size as a direct knob
for trading memory against preconditioner-update compute.

\subsection{\bluetext{Compute-time overhead}}\label{app:compute_overhead}
The additional cost of LLQR is concentrated in the periodic refitting of the
learned inverse preconditioner $\mU$, rather than in its application during the
outer optimization loop. Applying $\mU$ to the gradient is a structured matrix
action and does not require a linear solve. The main overhead therefore comes
from each preconditioner update, which requires one forward pass to cache the
per-layer JVP/VJP terms, one backward pass to cache the HVP terms, and a small
number of inner optimization steps to refit $\mU$.

In our ImageNet experiments, we use 25 inner forward/backward steps per
preconditioner update and perform this update only a few times per epoch
(typically 2--4 times). Under the current schedule, this corresponds to roughly
$10{,}400$ additional forward/backward steps over 100 epochs. By comparison,
ImageNet training with batch size 256 requires approximately $5{,}000$
forward/backward steps per epoch, or about $500{,}000$ steps over 100 epochs.
The resulting expected training-time multiplier is therefore approximately
\[
1 + \frac{10{,}400}{500{,}000} \simeq 1.02.
\]
Empirically, we measure a training-time multiplier of about $\times 1.03$ on
ImageNet. The small difference between the expected and measured values is
consistent with implementation overheads, and is likely due to the matrix-vector product of the preconditioner on the gradient made at every step.

\subsection{\bluetext{Block Structures}}\label{app:block_structures}
This subsection details the layerwise structures used to parameterize the learned inverse preconditioner $\mU$. In all cases, the structure is solely imposed on $\mU$, not on the divergence-induced metric before deriving the LQR objective. For kernel-like parameters, including dense, convolutional, embedding, and projection kernels, we reshape the corresponding gradient component into a matrix $\mX_i\in\mathbb{R}^{m_i\times n_i}$; bias and normalization-scale parameters are always treated with diagonal blocks.

\begin{itemize}
  \item \textbf{Diagonal.} The diagonal structure stores one scalar per parameter coordinate. For a vectorized gradient component $\vg_i$, the preconditioned component is
  \[
  \mU_i\vg_i = \vd_i\odot \vg_i,
  \]
  where $\vd_i$ is learned in the inner loop. This is the least expressive structure considered here, but it is memory efficient and provides a useful reference for purely coordinate-wise rescaling.

  \item \textbf{K-FAC.} The Kronecker-factored structure \citep{pmlr-v37-martens15} stores two learned factors for each matrix-shaped parameter component. If $\mX_i\in\mathbb{R}^{m_i\times n_i}$ denotes the reshaped gradient, the block action is
  \[
  \mY_i = \mC_i \mX_i \mD_i^\top,
  \qquad
  \operatorname{vec}(\mY_i)
  =
  (\mD_i\otimes \mC_i)\operatorname{vec}(\mX_i),
  \]
  with $\mC_i\in\mathbb{R}^{m_i\times m_i}$ and $\mD_i\in\mathbb{R}^{n_i\times n_i}$. Thus the inverse preconditioner block is represented through a Kronecker product rather than as a dense $m_i n_i\times m_i n_i$ matrix. Bias and scale components leverage the same diagonal structure as above.

  \item \textbf{E-KFAC.} The E-KFAC structure \citep{EKFAC--GeorgeLBBV18} separates a change of coordinates from the scaling applied in those coordinates. For a reshaped gradient $\mX_i$, it stores transform factors $\mQ_i^{\mathrm{L}}\in\mathbb{R}^{m_i\times m_i}$ and $\mQ_i^{\mathrm{R}}\in\mathbb{R}^{n_i\times n_i}$, together with a learned inverse diagonal $\vs_i\in\mathbb{R}^{m_i n_i}$. The operation is
  \[
  \widehat{\mX}_i
  =
  (\mQ_i^{\mathrm{L}})^\top \mX_i \mQ_i^{\mathrm{R}},
  \qquad
  \mY_i
  =
  \mQ_i^{\mathrm{L}}\,
  \operatorname{unvec}\!\left(
  \vs_i\odot \operatorname{vec}(\widehat{\mX}_i)
  \right)
  (\mQ_i^{\mathrm{R}})^\top,
  \]
  where $\operatorname{unvec}$ reshapes the scaled vector back to an $m_i\times n_i$ matrix. Equivalently,
  \[
  \operatorname{vec}(\mY_i)
  =
  (\mQ_i^{\mathrm{R}}\otimes \mQ_i^{\mathrm{L}})
  \operatorname{diag}(\vs_i)
  (\mQ_i^{\mathrm{R}}\otimes \mQ_i^{\mathrm{L}})^\top
  \operatorname{vec}(\mX_i).
  \]
  Bias and scale components again remain diagonal.
\end{itemize}

We note that the usefulness of the E-KFAC structure in LLQR should not be interpreted as requiring the learned factors to recover the exact curvature eigenbasis. In classical E-KFAC, the transform is tied to the eigenspaces of the Kronecker curvature factors, and the diagonal correction refines the approximation in that basis. In LLQR, where the inverse preconditioner is learned directly rather than constructed from an explicit curvature eigendecomposition, we believe E-KFAC structure should instead be viewed as an architectural bias. The factors $\mQ_i^{\mathrm{L}}$ and $\mQ_i^{\mathrm{R}}$ provide a structured change of coordinates for the layerwise gradient, while the diagonal vector $\vs_i$ learns anisotropic scaling in the transformed representation. This yields a more expressive inverse action than a pure Kronecker product, because individual transformed coordinates can be scaled separately, while preserving an efficient structured form.

\subsection{\mdfy{Ablation Findings}}
\begin{itemize}
  \item \textbf{Damping.} We set damping to zero in all experiments and omit it from the algorithm for clarity. Empirically, small damping values had no measurable effect, while large values degraded performance. We attribute this to (i) the absence of explicit matrix inversions, which reduces sensitivity to conditioning, and (ii) the limited number of inner-loop updates, which mitigates the impact of extreme eigenvalues.

  \item \textbf{Inner-loop optimization.} Performance depends on the interaction between update frequency, number of inner steps, and inner-loop learning rate. Less frequent updates or fewer inner steps can be compensated by scaling the inner loop learning rate. Since update frequency dominates computational cost, we recommend 1--4 updates per epoch with 25--50 inner steps. Adam and SGDM perform comparably as inner optimizers; we default to SGDM due to its lower memory and computational footprint.

  \item \textbf{$\mU$ update stabilization.} We use EMA to update $\mU$ at every inner loop. Larger EMA decay values yield more stable inner-loop updates. Setting $\beta \in [0.85-0.95]$ is a robust default choice. In practice, this approach facilitates the exploration of the inner loop learning rate.

  \item \textbf{Warm-starting $\mU$.} We evaluated warm-starting the inner-loop optimization of $\mU$ from its previous value. While this slightly accelerated very early convergence, it led to less stable training and weaker final performance, particularly on ImageNet, where results only matched SGDM. We also observed mild performance degradation after grokking in grokking-style experiments. In addition, warm-starting required careful tuning, including explicit inner-loop learning-rate decay. For simplicity and robustness, we therefore default to EMA-style updates, reinitializing $\mU$ to the identity at each inner-loop update.

  \item \textbf{Conjugate-gradient inner solvers.} Since the relaxed LLQR
  subproblem is quadratic in the preconditioner action, we also explored
  conjugate-gradient-based variants for updating $\mU$
  \citep{Hestenes1952MethodsOC}. The main motivation was to replace the
  additional inner-loop learning rate by a line-search-based solver. In
  practice, however, CG was not a clear improvement in our experiments. While
  the resulting updates were functional, they did not improve performance,
  introduced a noticeable computational slowdown, and remained sensitive to the
  number of inner iterations used for each preconditioner update. This tuning
  burden offset much of the practical benefit of removing the inner-loop
  learning rate. By contrast, SGD-like optimizers are mature in neural-network
  settings and align naturally with the layerwise relaxed LLQR objective, whose
  optimization resembles a small structured neural-network training problem.
  We therefore view CG as a viable alternative implementation choice, but not
  as a clearly superior default solver under our current setup.

\end{itemize}

\subsection[Sensitivity sweeps]{\crtext{Sensitivity sweeps}}
\label{app:sensitivity_sweeps}

\crtext{We evaluate LLQR with the NGD-induced quadratic and E-KFAC structure on ResNet-18/CIFAR--100, keeping the outer SGDM hyperparameters fixed. Each swept configuration uses three seeds; we reuse the nominal LLQR with five-seed as reference.}

\crtext{Over the tested ranges, the largest accuracy losses occur with large inner learning rates. Increasing the number of inner steps from $25$ to $50$--$250$ also lowers mean accuracy by $1.08$--$1.52$ percentage points. None of the tested Adam settings improves on the nominal inner SGDM setting. EMA decays of $0.75$--$0.99$ retain gains over unmodified SGDM; update periods of $100$--$1500$ training steps remain within $0.18$ points of the nominal LLQR mean, and damping values up to $10^{-2}$ remain within $0.34$ points.}

\par\addvspace{\medskipamount}
\noindent\begin{minipage}{\linewidth}
\captionsetup{type=table}
\caption{\crtext{Inner optimizer and learning-rate sweeps. Entries are mean accuracy differences (percentage points) $\pm$ standard errors relative to LLQR ($79.96\pm0.10\%$) or unmodified SGDM ($78.97\pm0.22\%$). Asterisks mark nominal settings; unvaried parameters remain fixed; reference self-differences are zero.}}
\label{tab:sensitivity_optimization}
\begin{subtable}[t]{0.49\linewidth}
\centering
\caption{\crtext{Inner optimizer and its learning rate}}
\small
\setlength{\tabcolsep}{3pt}
\crtext{\begin{tabular}{lcc}
\toprule
Optimizer / LR & $\Delta$ LLQR & $\Delta$ SGDM \\
\midrule
SGDM $5\!\times\!10^{-3}$$^{*}$ & $0.00\pm0.00$ & $+0.99\pm0.24$ \\
Adam $10^{-4}$ & $-0.95\pm0.18$ & $+0.04\pm0.26$ \\
Adam $5\!\times\!10^{-4}$ & $-0.41\pm0.17$ & $+0.59\pm0.26$ \\
Adam $10^{-3}$ & $-0.42\pm0.15$ & $+0.58\pm0.24$ \\
Adam $5\!\times\!10^{-3}$ & $-0.31\pm0.13$ & $+0.68\pm0.23$ \\
Adam $10^{-2}$ & $-0.54\pm0.26$ & $+0.45\pm0.33$ \\
Adam $5\!\times\!10^{-2}$ & $-7.94\pm1.32$ & $-6.95\pm1.33$ \\
\bottomrule
\end{tabular}}
\end{subtable}\hfill
\begin{subtable}[t]{0.49\linewidth}
\centering
\caption{\crtext{Inner learning rate (SGDM)}}
\small
\setlength{\tabcolsep}{3pt}
\crtext{\begin{tabular}{lcc}
\toprule
Learning rate & $\Delta$ LLQR & $\Delta$ SGDM \\
\midrule
$10^{-4}$ & $-0.80\pm0.24$ & $+0.19\pm0.31$ \\
$5\!\times\!10^{-4}$ & $-0.67\pm0.18$ & $+0.33\pm0.27$ \\
$10^{-3}$ & $-0.75\pm0.11$ & $+0.24\pm0.22$ \\
$5\!\times\!10^{-3}$$^{*}$ & $0.00\pm0.00$ & $+0.99\pm0.24$ \\
$10^{-2}$ & $-1.00\pm0.27$ & $-0.01\pm0.33$ \\
$5\!\times\!10^{-2}$ & $-17.50\pm4.70$ & $-16.50\pm4.70$ \\
\bottomrule
\end{tabular}}
\end{subtable}
\end{minipage}
\par\addvspace{\medskipamount}

\par\addvspace{\medskipamount}
\noindent\begin{minipage}{\linewidth}
\captionsetup{type=table}
\caption{\crtext{Sensitivity to inner steps and preconditioner update period. Reporting conventions and reference values follow Table~\ref{tab:sensitivity_optimization}.}}
\label{tab:sensitivity_schedule}
\begin{subtable}[t]{0.49\linewidth}
\centering
\caption{\crtext{Number of inner steps}}
\small
\setlength{\tabcolsep}{3pt}
\crtext{\begin{tabular}{lcc}
\toprule
Inner steps & $\Delta$ LLQR & $\Delta$ SGDM \\
\midrule
$5$ & $-0.67\pm0.13$ & $+0.32\pm0.23$ \\
$10$ & $-0.30\pm0.19$ & $+0.69\pm0.27$ \\
$25$$^{*}$ & $0.00\pm0.00$ & $+0.99\pm0.24$ \\
$50$ & $-1.08\pm0.18$ & $-0.09\pm0.26$ \\
$100$ & $-1.41\pm0.23$ & $-0.41\pm0.30$ \\
$250$ & $-1.52\pm0.30$ & $-0.52\pm0.35$ \\
\bottomrule
\end{tabular}}
\end{subtable}\hfill
\begin{subtable}[t]{0.49\linewidth}
\centering
\caption{\crtext{Preconditioner update period}}
\small
\setlength{\tabcolsep}{3pt}
\crtext{\begin{tabular}{lcc}
\toprule
Training steps & $\Delta$ LLQR & $\Delta$ SGDM \\
\midrule
$10$ & $-0.78\pm0.16$ & $+0.21\pm0.25$ \\
$50$ & $-0.36\pm0.13$ & $+0.63\pm0.23$ \\
$100$ & $-0.12\pm0.30$ & $+0.87\pm0.35$ \\
$250$ & $-0.18\pm0.24$ & $+0.81\pm0.31$ \\
$500$$^{*}$ & $0.00\pm0.00$ & $+0.99\pm0.24$ \\
$1000$ & $-0.06\pm0.19$ & $+0.93\pm0.27$ \\
$1500$ & $-0.14\pm0.19$ & $+0.86\pm0.27$ \\
$2500$ & $-0.30\pm0.24$ & $+0.70\pm0.31$ \\
\bottomrule
\end{tabular}}
\end{subtable}
\end{minipage}
\par\addvspace{\medskipamount}

\par\addvspace{\medskipamount}
\noindent\begin{minipage}{\linewidth}
\captionsetup{type=table}
\caption{\crtext{Sensitivity to EMA decay and damping. Reporting conventions and reference values follow Table~\ref{tab:sensitivity_optimization}.}}
\label{tab:sensitivity_stabilization}
\begin{subtable}[t]{0.49\linewidth}
\centering
\caption{\crtext{EMA decay}}
\small
\setlength{\tabcolsep}{3pt}
\crtext{\begin{tabular}{lcc}
\toprule
Decay & $\Delta$ LLQR & $\Delta$ SGDM \\
\midrule
$0.4$ & $-1.45\pm0.16$ & $-0.45\pm0.25$ \\
$0.5$ & $-1.39\pm0.26$ & $-0.40\pm0.32$ \\
$0.6$ & $-0.82\pm0.20$ & $+0.18\pm0.28$ \\
$0.75$ & $-0.40\pm0.18$ & $+0.60\pm0.26$ \\
$0.8$ & $-0.34\pm0.11$ & $+0.65\pm0.22$ \\
$0.85$ & $-0.23\pm0.18$ & $+0.77\pm0.26$ \\
$0.9$ & $-0.19\pm0.14$ & $+0.80\pm0.24$ \\
$0.95$$^{*}$ & $0.00\pm0.00$ & $+0.99\pm0.24$ \\
$0.99$ & $-0.32\pm0.19$ & $+0.67\pm0.27$ \\
\bottomrule
\end{tabular}}
\end{subtable}\hfill
\begin{subtable}[t]{0.49\linewidth}
\centering
\caption{\crtext{Damping}}
\small
\setlength{\tabcolsep}{3pt}
\crtext{\begin{tabular}{lcc}
\toprule
Damping & $\Delta$ LLQR & $\Delta$ SGDM \\
\midrule
$0$$^{*}$ & $0.00\pm0.00$ & $+0.99\pm0.24$ \\
$10^{-6}$ & $-0.20\pm0.24$ & $+0.79\pm0.31$ \\
$10^{-5}$ & $-0.20\pm0.20$ & $+0.79\pm0.28$ \\
$10^{-4}$ & $-0.22\pm0.13$ & $+0.78\pm0.23$ \\
$10^{-3}$ & $-0.34\pm0.19$ & $+0.65\pm0.27$ \\
$10^{-2}$ & $-0.22\pm0.16$ & $+0.77\pm0.25$ \\
\bottomrule
\end{tabular}}
\end{subtable}
\end{minipage}
\par\addvspace{\medskipamount}


\section{Experiment Details}\label{app:experiment_details}

\subsection{Rosenbrock Validation}\label{app:rosenbrock_details}

\begin{wrapfigure}{l}{0.50\linewidth}
  \centering
  \vspace{-1.5em}
  \includegraphics[width=1.0\linewidth]{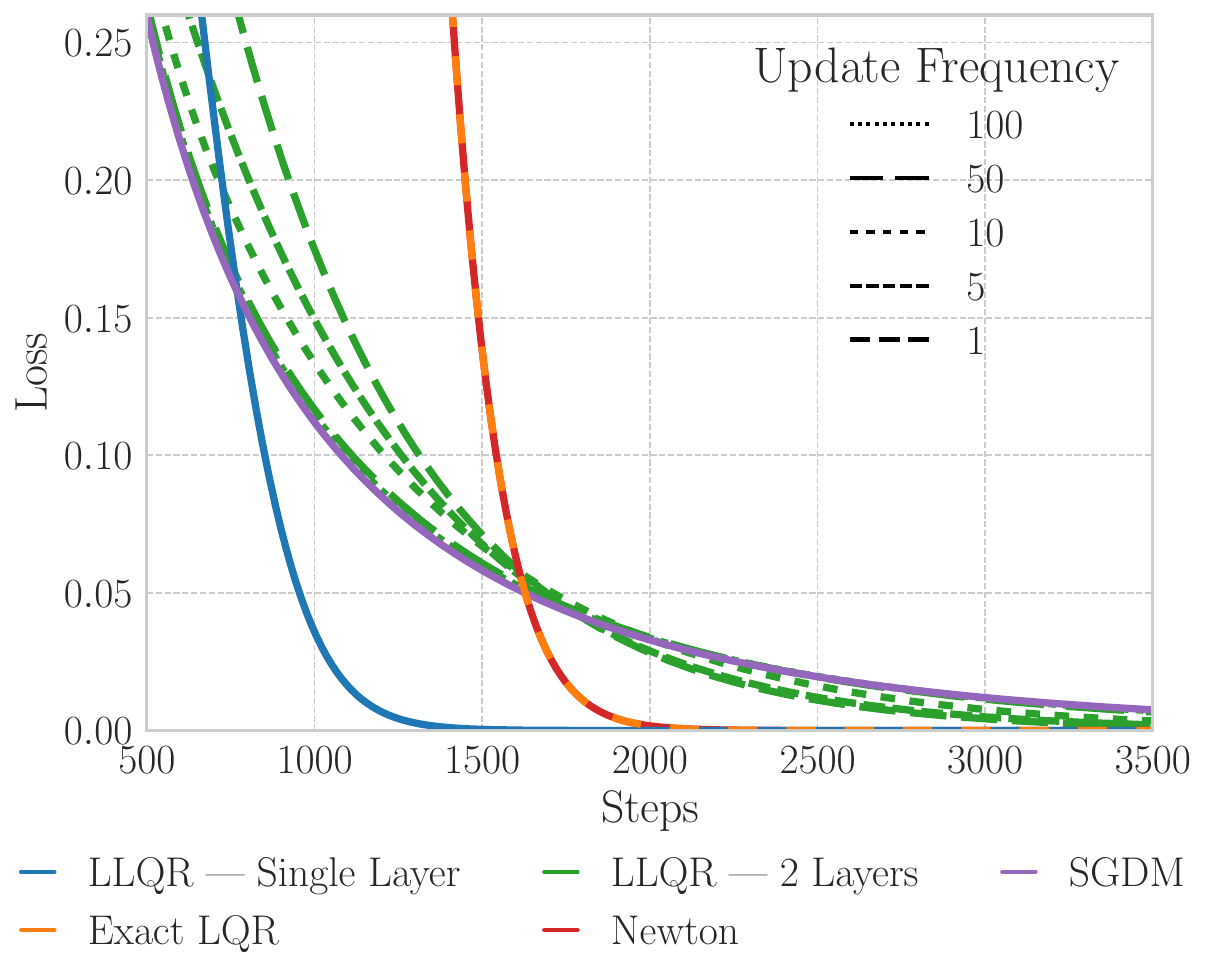}
  \caption{Validation of the LQR equivalence and relaxation on the Rosenbrock function.
(\textcolor{red}{\textbf{Red}} and \textcolor{orange}{\textbf{Orange}}) Newton’s step and Riccati solution overlap, confirming exact equivalence.
(\textcolor{green}{\textbf{Green}}) Relaxation with block–diagonal $\mU$ converges faster as update frequency increases, though remaining approximate.
(\textcolor{blue}{\textbf{Blue}}) Relaxation with dense $\mU$ on the single-layer formulation matches the convergence rate of Newton’s step.
(\textcolor{violet}{\textbf{Violet}}) SGDM is provided for comparison. \mdfy{As no line search is used during optimization, the comparison focuses solely on convergence rates.}}
  \label{fig:rosenbrock}
  \vspace{-3.5em} 
\end{wrapfigure}

The two-dimensional Rosenbrock function is defined as:
\begin{equation}
R_{a,b}(x,y)=
(a - x)^2 \;+\; b\,\big(y - x^2\big)^2
\end{equation}
This can equivalently be reformulated as a two-layer function:
\begin{equation}
    \begin{aligned}
        &\vx_1 = [\,u_1,\,u_2,\,u_3\,]^\top = [\,a^2 - 2ax,\; x^2,\; b\,]^\top\\
        &R_{a,b}(x,y)=(u_1 + u_2) \;+\; u_3\,\big(y - u_2\big)^2
    \end{aligned}
\end{equation}

In our validation setup, we optimize for the minimum of $R_{1,100}$ with respect to $x,y$. In the relaxed setup, the inverse preconditioner $\mU$ is learned using SGDM with an inner loop learning rate of $10^{-4}$ over 500 inner steps at every update of the preconditioner. 
Results are shown in Fig.~\ref{fig:rosenbrock}.

\subsection{ResNet-18 Experiments}\label{app:CIFAR_experiments_details}
\paragraph{CIFAR-100: }
We trained ResNet-18 models for 200 epochs, using a batch size of 128 and a weight decay of $10^{-3}$.
A cosine annealing schedule was applied to gradually reduce the learning rate over training.
For SGDM, we used an initial learning rate of $0.05$ and a momentum of $0.9$. For AdamW, the initial learning rate was rather $5\times10^{-4}$.
All experiments were conducted from scratch with identical random seeds across optimizers to ensure fair comparison. We applied standard data augmentation techniques widely used in CIFAR training: random horizontal flips, random crops with 4-pixel padding, normalization with dataset statistics, and Cutout regularization \citep{devries2017cutout}.

\paragraph{CIFAR-10: }
The setting was the same as for CIFAR-100 runs, except for the weight decay of $5\times10^{-4}$ and AdamW initial learning rate set to $10^{-3}$.

\paragraph{AdaFisher setup on CIFAR--100: } \label{app:adafisher_training_details}
Most settings remain the same---including data augmentation---except for the weight decay of $5\times10^{-4}$, an initial learning rate set to $0.1$ and a batch size of 256.

\subsection{ResNet-50 on ImageNet} \label{app:imagenet_details}
\mdfy{We trained ResNet-50 models on ImageNet for 100 epochs using a batch size of 256 and a weight decay of $10^{-4}$. Optimization was performed with SGDM using an initial learning rate of $0.1$ and a momentum of $0.9$. A warmup followed by a piecewise decay learning-rate schedule was applied throughout training. Label smoothing with coefficient $0.1$ was used. All experiments were trained from scratch with fixed initialization and random seeds to ensure fair comparison across optimizers. For data augmentation, we applied standard ImageNet preprocessing: random $224\times224$ crops from resized $256\times256$ images and random horizontal flips during training, and central cropping at evaluation.}

For Fig.~\ref{fig:imagenet_and_iwslt14_training_curves}, curves are averaged over five seeds, shaded regions denote standard errors. All ImageNet experiments were run on NVIDIA A100 GPUs.

\subsection[Learning the inverse preconditioner]{Learning $\mU$: }
\mdfy{Across all experiments, we learned the inverse preconditioner with 25 steps of SGDM (constant learning rate of $0.005$, momentum of $0.9$) at every 500 CIFAR iterations (250 iterations in AdaFisher setup with $\mathrm{BS}=256$ to keep update ratio per epoch constant) and 1500 ImageNet iterations of the main optimization loop. EMA---with $\beta$ set to $0.9$ across all experiments approximating Newton's descent and to $0.95$ for NGD---was used to update $\mU$. Damping was kept to 0.}

\subsection{Grokking Experiments}\label{app:grokking_experiments_details}
Grokking experiments were conducted on 5 standard algorithmic datasets (addition, subtraction, multiplication, division, exponentiation), all modulo prime number $p=97$~\citep{DBLP:journals/corr/abs-2201-02177} and using 60\% of the dataset for training (40\% for test). Models were trained from scratch with a batch size of $512$ and a fixed learning rate of $10^{-3}$.
AdamW was used, with $\beta_2 = 0.98$, momentum of $0.9$, and no weight decay to further extend the overfitting plateau \citep{DBLP:journals/corr/abs-2201-02177}.
Training ran until until grokking occurred, i.e. when top-1 test accuracy reached 95\% after initial overfitting.

When learning $\mU$, most hyperparameters were kept as above, except for the learning rate for the diagonal structure, which was set to $0.01$, and the EMA decay, which was lowered to values in $[0.65, 0.8]$ (per dataset) to enable faster grokking. Damping was kept at 0.

For Fig.~\ref{fig:grokking-boxplot}, results are displayed across 10 random seeds; boxes show the interquartile range (Q1–Q3) with the center line being the median. Whiskers extend to the most extreme points within 1.5×IQR, and points beyond the whiskers are outliers. All experiments were run on NVIDIA L40S GPUs.

\subsection{IWSLT14 German to English}\label{app:iwslt14_details}
\mdfy{We trained a Fairseq-style Transformer on IWSLT14 German-to-English translation using AdamW with learning rate $5\times10^{-4}$, betas $(0.9,0.98)$, weight decay $10^{-4}$, dropout $0.3$, label smoothing $0.1$, and an inverse-square-root learning-rate schedule. Data were prepared with the standard Fairseq IWSLT14 De--En preprocessing pipeline, using Moses tokenization, lowercasing, a shared 10k BPE code, and separate source and target dictionaries. BLEU was evaluated with beam size $5$, length rule $1.2\cdot |\mathrm{src}|+10$, BPE removal, and Moses detokenization.}

\mdfy{For LLQR, we used an E-KFAC block structure with the NGD-induced divergence and learned $\mU$ with the same inner optimizer as above: 25 SGDM steps, inner learning rate $0.005$, momentum $0.9$, and damping $0$. The preconditioner was updated every 250 training iterations. The only LLQR hyperparameter changed relative to the preceding NGD experiments was the EMA decay for $\mU$, set to $0.925$ instead of $0.95$. All results are averaged over five seeds and the reported time multiplier was measured on NVIDIA L40S GPUs.}

\subsection[Autoregressive Language Modeling]{\crtext{Autoregressive Language Modeling}}
\label{app:wikitext103_details}

\crtext{We trained a decoder following the Fairseq GPT-3 Small architecture on WikiText-103 with its full word-level vocabulary, giving approximately 291M parameters. The model has 12 pre-normalized Transformer blocks, hidden dimension 768, feed-forward dimension 3072, 12 attention heads, learned positional embeddings, and tied input embeddings and output projection. Dropout and label smoothing were set to zero. Training used sequences of 512 tokens, microbatches of four sequences, and 32 gradient-accumulation steps, giving 65,536 nominal training target tokens per update. We report all methods through 15,000 updates, approximately 0.98 billion nominal training target tokens. Each run used one NVIDIA L40S GPU, BF16 computation, and FP32 parameters and optimizer state. We used five matched initialization seeds, with the same data order across methods.}

\crtext{Both base optimizers used an inverse-square-root learning-rate schedule with 4,000 warmup updates, an initial learning rate of $10^{-7}$, a peak learning rate of $5\times10^{-4}$, and decoupled weight decay $0.01$. AdamW used betas $(0.9,0.98)$. For Muon, the hidden attention and MLP matrices used Nesterov momentum $0.95$, five Newton--Schulz iterations, and the per-matrix update scaling $0.2\sqrt{\max(m,n)}$ for an $m\times n$ matrix, following the scaled-Muon recipe of \citet{DBLP:journals/corr/abs-2502-16982}. Token and positional embeddings, the tied output head, normalization parameters, and biases used auxiliary AdamW with the same betas. The learning-rate schedule and weight decay were shared across both partitions and kept unchanged when adding LLQR.}

\crtext{LLQR used the NGD-induced quadratic and the E-KFAC structure of Appendix~\ref{app:block_structures}, with a diagonal vocabulary-side factor and a dense hidden-side factor for the tied embedding/output matrix. This avoids storing a quadratic-size vocabulary factor. We fitted $\mU$ every 200 updates using 25 SGDM steps with learning rate $0.005$, momentum $0.9$, zero damping, and an inner minibatch of one 512-token sequence. LLQR settings were selected separately for the two base optimizers using validation perplexity: the EMA decay was $0.6$ for AdamW and $0.4$ for Muon. LLQR preconditions gradients before the base optimizer's moment updates, including before Muon's momentum and orthogonalization. In the auxiliary-only control, we used the same LLQR fitting settings as for Muon with LLQR on all parameters, but passed preconditioned gradients only to auxiliary AdamW and left the gradients entering Muon unchanged.}

\crtext{Validation was evaluated every 1,000 updates. For each configuration, Table~\ref{tab:wikitext103_llqr} reports the lowest five-seed mean validation perplexity over this reporting window, together with token accuracy at the same checkpoint; uncertainties are standard errors across seeds. These are best-checkpoint results rather than final-checkpoint values. The threshold columns identify the first evaluated checkpoint with mean perplexity below $30$, without interpolation. Elapsed times are averaged across seeds at the corresponding update count and include compilation and preceding evaluations within the recorded training loop. The fixed-update time ratios in Section~\ref{sec:autoregressive_lm} compare each LLQR variant with its own base optimizer.}

\end{document}

%% file: DL_notation.tex
\usepackage{bm}

\def\mA{{\bm{A}}}
\def\mB{{\bm{B}}}
\def\mC{{\bm{C}}}
\def\mD{{\bm{D}}}

\def\mF{{\bm{F}}}

\def\mH{{\bm{H}}}
\def\mI{{\bm{I}}}

\def\mK{{\bm{K}}}

\def\mM{{\bm{M}}}

\def\mP{{\bm{P}}}
\def\mQ{{\bm{Q}}}
\def\mR{{\bm{R}}}

\def\mU{{\bm{U}}}

\def\mW{{\bm{W}}}
\def\mX{{\bm{X}}}
\def\mY{{\bm{Y}}}

\def\vtheta{{\bm{\theta}}}
\def\vlambda{{\bm{\lambda}}}
\def\va{{\bm{a}}}

\def\vd{{\bm{d}}}

\def\vg{{\bm{g}}}

\def\vp{{\bm{p}}}
\def\vq{{\bm{q}}}
\def\vr{{\bm{r}}}
\def\vs{{\bm{s}}}

\def\vu{{\bm{u}}}

\def\vw{{\bm{w}}}
\def\vx{{\bm{x}}}
\def\vy{{\bm{y}}}
\def\vz{{\bm{z}}}

\newcommand{\KL}{D_{\mathrm{KL}}}

\DeclareMathOperator*{\argmin}{arg\,min}

\renewcommand{\vec}[1]{\boldsymbol{#1}}

\newcommand{\x}{\vec{x}}

\renewcommand{\vtheta}{\vec{\theta}}
\newcommand{\vlam}{\boldsymbol{\lambda}}

